\documentclass[a4paper, 11pt]{article}
\makeatletter
\renewcommand\paragraph{\@startsection{paragraph}{4}{\z@}%
	{-3.25ex\@plus -1ex \@minus -.2ex}%
	{1.5ex \@plus .2ex}%
	{\normalfont\normalsize\bfseries}}
\makeatother

\usepackage[utf8]{inputenc}
\usepackage{amssymb}
\usepackage{amsmath}
\usepackage{mathtools}
\usepackage[Algorithmus]{algorithm}
\usepackage{algpseudocode}
\usepackage{graphicx}
\usepackage[english]{babel}   
\usepackage[T1]{fontenc}
\usepackage[nonumberlist,acronym,toc]{glossaries}
\usepackage{url}
\usepackage{multirow}
\usepackage{array}

\usepackage{tabulary}
\usepackage{tabularx}
\usepackage{csquotes}
\usepackage{longtable}
\usepackage[onehalfspacing]{setspace}

\usepackage{listings}

\usepackage{float}
\usepackage{caption}

\title{Method for the semantic indexing of concept hierarchies, uniform representation, use of relational database systems and generic and case-based reasoning}
\usepackage{authblk}
\author[*]{Uwe Petersohn}
\author[*]{Sandra Zimmer}
\author[**]{Jens Lehmann}

\affil[*]{\footnotesize{TU Dresden, Faculty of Computer Science, Institute for Artificial Intelligence, D-01062 Dresden, Germany}}
\affil[**]{\footnotesize{Frauenhofer Institut für Intelligente Analyse und Informationssysteme IAIS, D-53115 Bonn and D-01069 Dresden, Germany}}

\date{\today}

\usepackage{palatino,graphicx}  
\usepackage{amsthm}
\newtheoremstyle{break}
{\topsep}{\topsep}
{\itshape}{}
{\bfseries}{}
{\newline}{}
\theoremstyle{break}
\newtheorem{Def}{Definition}
\newtheorem{Bsp}{Example}
\newtheorem{Alg}{Algorithm}
\newtheorem{Satz}{Proposition}
\newtheoremstyle{break}
{\topsep}{\topsep}
{\itshape}{}
{\bfseries}{}
{\newline}{}
\theoremstyle{break}

\newcommand{\key}{X}
\newcommand{\dconc}{K^{C}}
\newcommand{\dconcprime}{K^{C'}}
\newcommand{\dconx}{K}
\newcommand{\dconset}{KM^{C}}
\newcommand{\bset}{M}
\newcommand{\gkey}{K}

\usepackage{fancyhdr}
\begin{document}
\def\autor{\sf U.Petersohn, S. Zimmer \\ J. Lehmann}
\def\institut{\sf Institut f\"ur K\"unstliche Intelligenz}

\maketitle

\begin{abstract}
	This paper presents a method for semantic indexing and describes its application in the field of knowledge representation.
	Starting point of the semantic indexing is the knowledge represented by concept hierarchies.
	The goal is to assign keys to nodes (concepts) that are hierarchically ordered and syntactically and semantically correct.
	With the indexing algorithm, keys are computed such that concepts are partially unifiable with all more specific concepts and only semantically correct concepts are allowed to be added.
	The keys represent terminological relationships.
	Correctness and completeness of the underlying indexing algorithm are proven.
	The use of classical relational databases for the storage of instances is described.
	Because of the uniform representation, inference can be done using case-based reasoning and generic problem solving methods.
\end{abstract}

\section{Introduction}
\paragraph{Problem statement}
Modern methods of knowledge representation, like description logic (e.g. \cite{Baad03}, \cite{Baad11}) and ontologies (e.g. \cite{StSt13}), fulfill the properties of formal semantics and high expressiveness.
Furthermore, they enable powerful inference procedures and are suitable for a wide range of applications.
Various capable development environments and software tools exist.
\par
The main aim of this paper is the development of a methodology that, on one hand, permits the conceptual recording and structuring of the application domain through concepts, concept hierarchies, multi-axial composition of concepts and concept descriptions.
On the other hand, it uses a representation that allows for dynamic dialog systems, case-based and generic problem solving methods and their interactions with relational databases to be implemented in one system.
This modeling is, for example, relevant in medical domains, when solutions to problems require the inclusion of knowledge from experience (treated cases) in addition to generic knowledge (textbook knowledge).

\paragraph{Representational paradigm}
A uniform and structured representation of the domain's concepts and concept hierarchies as well as the conceptual knowledge of the domain of discourse is achieved through a mapping to \enquote{semantic indices} (computed keys).
A partial unification of keys connects concepts of the domain, concept hierarchies and the defining concept descriptions and instances.
\par
Starting point of the semantic indexing are the concept hierarchies of the represented domain of discourse.
The goal of the semantic indexing is assigning a key to each node and each concept, with the key of a certain concept summarizing all keys for nodes of that concept (Chapter \ref{subsectionGrundbegriffe}).
The keys are computed with the presented algorithm \enquote{semantic indexing} (Chapter \ref{sct:MainIndexingAlgorithm}) and represent terminological relationships.
Correctness and completeness of the underlying indexing algorithm are proven.
Each key also contains its inheritance path.
Through a multiaxial modeling of concept hierarchies, a clear description is possible even for complex concepts, situations and expressions (Chapter \ref{subsct:multiaxial}).
The integrative application of different inference methods is possible (Chapters \ref{sct:BegriffshierarchienKonz}, \ref{section:ProblemSolving}).
\paragraph{Databases}
Knowledge processing systems often store their data in very different ways and are frequently focused on specific data structures for an efficient access.
The approach in this paper enables a uniform representation that permits the modeling and representation of practical domains as well as storage of instances in databases (Chapter \ref{sct:DB}), but also ensures an efficient analysis of the represented data and knowledge.
It connects the fields of knowledge representation and relational databases.
\par
The use of database systems for the maintenance of structures and instances with a uniform representation allows the implementation of efficient possibilities for accessing the stored data and knowledge as well as the usage of the extensive possibilities of modern database systems.
\paragraph{Applications}
As a result of the uniform knowledge representation, the setup, architecture and implementation of knowledge bases can be improved.
This is primarily done through the structured storage of knowledge from concept hierarchies, the storage of semantically clearly defined instances in a knowledge base as well as an efficient retrieval.
Different problem solving methods can be applied, especially in combination with each other, to evaluate the knowledge under different aspects, for example using logic-based, concept-based and case-based reasoning (Chapter \ref{section:ProblemSolving}).
\paragraph{Related Work}
Various different concept languages for terminological logics with different syntax, expressiveness and complexity exist, e.g.
KRIS\footnote{KRIS \textit{Knowledge Representation and Inference System} is a terminological logic with high expressiveness that was developed at the DFKI (German Research Center for Artificial Intelligence) \cite{KRIS91}.},
LOOM\footnote{LOOM is a terminological logic with very high expressiveness that was developed at the University of California. PowerLOOM, as the successor of LOOM, is even more expressive and uses a variant of the language KIF (\textit{Knowledge Interchange Format}) \cite{Loom07}.},
KRYPTON\footnote{KRYPTON combines frame-based representational language and theorem provers for the first order predicate logic \cite{KRYP85}.}
or
\par
ALCQUI
\footnote{ALCQI is based on the standard description logic ALC. It is extended by qualifying number restrictions and converse roles \cite{ALQI99}.}.
Other important languages are CRACK, KANDOR, SHF, NIKL, and SHIQ.
Similarly there exist different inference systems (e.g
FaCT\footnote{FaCT is a system that is based on tableau methods and supports OWL (\textit{web ontology language}) and OWL 2 \cite{FaCT19}.},
RACER\footnote{RACER (Renamed ABox and Concept Expression Reason) is a system of knowledge representation that was developed at the University of Hamburg, based on the very expressive logic SHIQ. It implements a strongly optimized ABox-tableau method \cite{RACER19}.})
and development environments (e.g. OntoSaurus\footnote{OntoSaurus is a graphical web browser for knowledge bases that are based on LOOM \cite{OntoS19}.}).
For the computation of the subsumption there are essentially two methods: NC-algorithms and tableau methods.
NC-algorithms are based on the syntactical comparison of the deduced concept descriptions.
The proof of correctness for the algorithm is simple, but the majority of NC-algorithms are incomplete.
Tableau methods, like they are implemented in e.g. KRIS, are theorem provers with \textit{backward chaining}.
Determining the subsumption of deduced concepts is turned into determining a contradiction and can be answered with conventional problem solving methods.
Correctness and completeness are provable.
Terminological reasoning is decidable but NP-complete.
Computing the subsumption for complex applications is often connected to an exponential time complexity.
The following compromises are possible:
\begin{itemize}
	\item Forgoing negation, disjunction and quantification respectively to reduce the expressiveness and achieve polynomial time complexity (KRYPTON, CLASSIC\footnote{CLASSIC was developed by AT\&T Laboratories and has implementations in Lisp, C and C++. It is one of the less expressive systems \cite{CLASS94}.})
	\item Forgoing completeness and decidability of inferences (NIKL, LOOM, KL-ONE\footnote{KL-ONE represents, as a representational language, the origin of terminological logics. KL-ONE is based on the formalization and generalization of the principles of frames and semantic networks.
	It serves the construction of complex structured concept descriptions \cite{KLON85}.})
	\item Accepting the exponential computing time (KRIS)
\end{itemize}

However, correct and complete inferences with polynomial time complexity are only possible with severely limited terminological logics.
\par
For ontologies, various different languages are available as well. Either informal graphical (e.g.
CML\footnote{The \textit{Conceptual Modelling Language} describes non-functional requirements of applications that were implemented in the programming language C \cite{CML94}.})
or formal description languages (e.g.
Ontolingua\footnote{Ontolingua is based on KIF \textit{(Knowledge Interchange Format)} and the \textit{frame ontology} and was developed especially for a formal specification of ontologies. KIF is a description language that was built on predicate logic and extended with language primitives. Thus meta-statements about relations can be made \cite{Grub92}.},
 CycL\footnote{CycL is the language on which the knowledge representation system CYC is based \cite{Cyc19}.},
 FLogic\footnote{FLogic is an integration of frame-based languages with the predicate calculus, which contains object oriented approaches \cite{KiLa89}.},
 RDFS\footnote{RDF-Schema is a W3C standard language for the XML-based representation of ontologies \cite{RDFS19}.},
 OIL\footnote{OIL is a web-based representation and inference layer for ontologies which builds on RDFS and expands its expressiveness \cite{OIL19}.},
DAML+OIL\footnote{DAML+OIL was developed for the realization of the semantic web. It is an independent continuation of OIL but the development has not continued since 2001, because OWL is held as the successor\cite{DAMLO19}.}) can be used.
In addition, there are development environments like OntoStudio\footnote{OntoStudio is a ontology development environment for creating and modifying ontologies. It supports among others the languages OWL and RDFS \cite{Onto19}.} or OntoSaurus, which support the creation of ontologies.
For inferences, ontology definitions are mapped to concrete operationalizations of the \textit{logical and operationalizational layer}, which possess formal semantics and enable correct and complete inferences.
Depending on the language used in the \textit{representation vocabulary layer}, the mapping happens in the \textit{logical and operationalizational layer}.
Depending on the chosen operationalization, the subsumption (exponential time complexity) or the predicate logic theorem prover (polynomial to exponential time complexity) can be applied as inference methods.
\par
Newer inference tools are being developed on the basis of the description languages. However, they are mostly based on the mentioned established languages.
For instance
in \cite{JLWO03} an XR-tree (XML Region Tree), a dynamic external memory index structure for strictly nested XML data, is proposed for retrieval. For a given element all its ancestors and / or descendants can be efficiently identified in an element set which is indexed by an XR-tree. The new devised stack-based structural join algorithm named XR-stack is able to evaluate two XR-tree indexed element sets regarding their structural relationship. This is done by avoiding unnecessary element scans by skipping ancestors and descendants which do not have matches. Experiments were conducted to show that the performance of XR-stack in comparison with the current state of the art significantly outperforms previous algorithms \cite{JLWO03}. 

Another approach is the use of LiteMat, an inference encoding scheme for large
RDF\footnote{RDF (\textit{Resource Description Framework }) is a standard developed by the W3C that forms the basis for the semantic web. Every expression consists of the 3-tuple subject, predicate, object \cite{RDF14}.}
 graphs, which is presented in \cite{CXNC19}.
Inferences are based on RDFS and the \texttt{owl:sameAs} property. The proposed extensions by integrating \texttt{owl:transitiveProperty} and \texttt{owl:inverseOf} properties enable to reach RDFS++ expressiveness. This has been achieved by assigning meaningful identifiers to elements of the TBox and ABox. This is efficient regarding memory space, query processing and speed of encoding \cite{CXNC19}.

\section{Basic concepts and introductory examples}
\label{subsectionGrundbegriffe}

The following chapters will introduce the representational paradigm and illustrate it with examples.

\subsection{Concept hierarchies}
\label{subsct:Begriffshierarchien}
 
\begin{Def}[Concept hierarchy] 
	\label{Begriffshierarchie}
	A concept hierarchy is a tree whose elements are concepts with the following properties:
	\begin{enumerate} 
		\item A node must not have more than one child node with the same concept.
		\item The dependency graph for the concept hierarchy is free of cycles.
	\end{enumerate}
\end{Def}

The first property means that it is not possible for multiple nodes with the same concept to belong to the same parent node.
A hierarchy is supposed to establish an order between its elements. Therefore, such a repetition would not be meaningful. In other words: Two identical child nodes being attached to the same parent node would contain the same information as only one of them being attached.
\par
\begin{figure}[!h]
	\centering
	\includegraphics[width=1.00\textwidth]{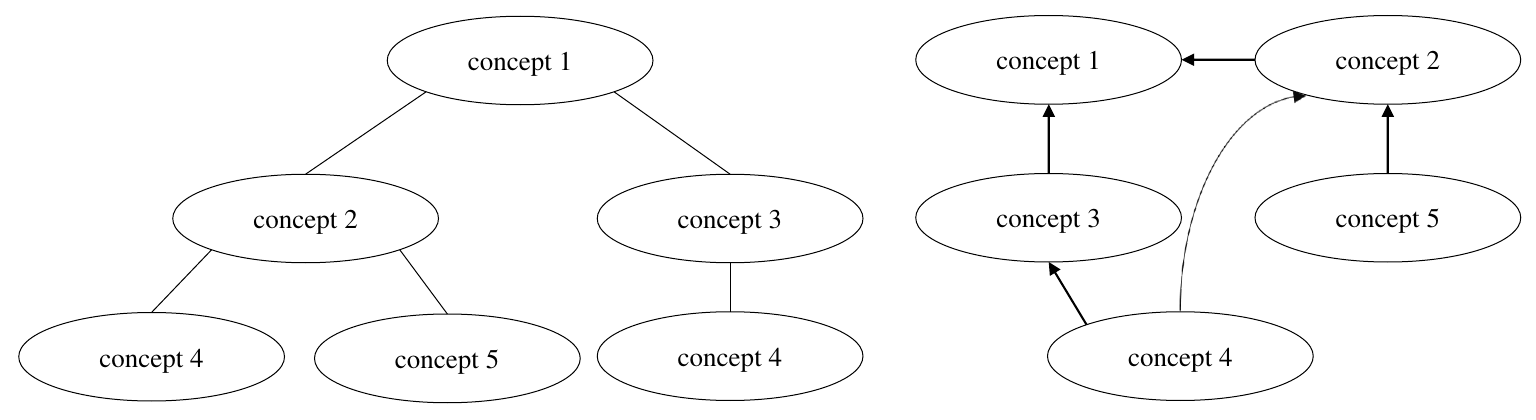}
	\caption[Simple concept hierarchy and the respective dependency graph.]{Simple concept hierarchy and the respective dependency graph.}
	\label{Abb:Abhaengigkeitsgraph}
\end{figure}

Figure \ref{Abb:Abhaengigkeitsgraph} shows an example of a concept hierarchy.
It is a tree whose nodes are concepts. Each concept can appear multiple times.

A dependency graph can be created for every hierarchy (Figure \ref{Abb:Abhaengigkeitsgraph}, on the right).
It contains all occurring concepts as nodes.
In the dependency graph, an edge points from \enquote{node A} to \enquote{node B} if \enquote{node A} is at least once in the concept hierarchy directly below \enquote{node B}.
In the example, \enquote{concept 2} and \enquote{concept 3} are directly below the root of the concept hierarchy \enquote{concept 1}. Therefore, edges from \enquote{concept 2} and \enquote{concept 3} point to \enquote{concept 1} in the dependency graph.
\par
To explain the second part of Definition \ref{Begriffshierarchie}, a few more terms have to be introduced.
\par
A circuit in a graph is a walk in which start and end node are the same.
A circuit $K_{1} \rightarrow ... \rightarrow K_{n}$ in a graph is a cycle if $K_{1} \rightarrow ... \rightarrow K_{n}$ is a path.
The absence of cycles has to be proven.

\begin{Def}[Path]
	A path $K_{1} \rightarrow ... \rightarrow K_{n}$ in the concept hierarchy is a sequence of nodes in which all $K_{i}$ with $1\leq i \leq n$ are nodes in the hierarchy and all $K_{i+1}$ with $1 \leq i \leq n-1$ are directly below $K_{i}$ in the hierarchy. 
\end{Def}

\begin{Def}[More specific, more general]
A concept $A$ is more specific than a concept $B$ in a certain concept hierarchy if there is a path in the dependency graph that leads from $A$ to $B$.
Conversely, a concept $A$ is more general than a concept $B$ in a certain concept hierarchy if there is a path in the dependency graph that leads from $B$ to $A$.
	\label{Def:allgspezieller}
\end{Def}

Another important feature of semantic indexing are the keys.
For this, some concepts have to be formally defined.

\begin{Def}[Keys]
A key $\key=\left[a_{1}, ..., a_{m}\right]$ is a comma-separated list. Every element $a_{i}$ with $1\leq i \leq m$ of this list is either a constant or a variable x.
\label{DefSchlüssel}
\end{Def}

The concept hierarchy has only one variable $x$. However, it can appear at multiple positions in the list.
The different $x$ variables are completely independent from each other.
\begin{Def}[Length of a key] 
	The length of a key is the number of elements of that key.
\end{Def}

\begin{Def}[Partial instance]
A key $\key_{1}=\left[a_{1}, ..., a_{m}\right]$ of length $m$ is a partial instance of a key $\key_{2}=\left[b_{1}, ..., b_{n}\right]$ of length $n$ if at least one variable $a_i$ is substituted by a constant $b_i$: $a_{i}=b_{i}$ with $0 \leq i \leq m$ and $b_{i}\neq x$.
\end{Def}

\begin{Def}[Instance]
A key $\key_{1}=\left[a_{1}, ..., a_{m}\right]$ of length $m$ is an instance of a key  $\key_{2}=\left[b_{1}, ..., b_{n}\right]$ of length $n$ if variables $a_i$ get substituted by constants $b_i$ and $n=m$,  $a_{i}=b_{i}$ for all $i$ with $0 \leq i \leq n$ and $b_{i}\neq x$. 
\end{Def}

\begin{Def}[Set of all instances] 
 For a key $\key$, $inst(\key)$ is the set of all instances of $\key$.
\end{Def}

Instances of a key can be represented by substituting the variables $x$ at different positions with constants.
It is not necessary to substitute all variables.
\par
Furthermore, a hierarchy of keys has to be defined for this context as well.
Keys can be more specific than other keys in two ways: They can be longer or they can substitute variables with constant symbols.

\begin{Def}[Initial key]
For a key $\key=\left[a_{1}, ..., a_{n}\right]$ and $1 \leq m \leq n$, $\key\left[0,m\right]$ denotes the initial key ${\key}'=\left[a_{1}, ..., a_{m}\right]$.
\end{Def}

\begin{Def}[Position within a key]
	For a key  $\key=\left[a_{1}, ..., a_{n}\right]$ and $1 \leq m \leq n$, $\key\left[m\right]$ denotes $a_{m}$, the $m$-th position in $\key$.
\end{Def}

\begin{Def}[Partial unification]
	A key $\key_{1}$ is partially unifiable with a key $\key_{2}$ if there exists an instance of $\key_{1}$ that is initial key of an instance of $\key_{2}$.
	\label{Def_unifizierbar}
\end{Def}

\par
Partial unifiability will now be explained with an example.
\begin{Bsp}[Partially unifiable]
$\key_{1}=\left[0, x, 2, x, 5\right]$ is partially unifiable with $\key_{2}=\left[0, 3, x, x, 5, x, 1\right]$, because the instance $\left[0, 3, 2, x, 5\right]$ of $\key_{1}$ is an initial key of the instance $\left[0, 3, 2, x, 5, x, 1\right]$ of $\key_{2}$.
\end{Bsp}
An indexing algorithm has to compute the keys such that a concept is partially unifiable with all more specific concepts and only semantically correct concepts can be inserted.
In the database, only the instances are represented. The complete trees are not stored.
This allows efficient access and the usage of keys for retrieval, inferences and other processes.
\par
The indexing algorithm creates node and concept keys.

\begin{Def}[Node keys, concept keys]
The key of a node $K$ is the node key $\key_{K}$. The key of a concept $B$ is the concept key $\underline{\key}_{B}$ (concept keys are underlined).
	\label{DefKnotenBegriff}
\end{Def}
This distinction is necessary because concepts can appear multiple times in the concept hierarchy.
A concept key has to correctly index its concept in all places of the concept hierarchy.
The node key represents the instances, including the inheritance path, within the database.

\begin{Bsp}[Node keys, concept keys]
The following table compares node and concept keys.
\setlength\LTleft{0pt}
\setlength\LTright{0pt}
\begin{center}
	\begin{scriptsize}
		\begin{minipage}{0.75\textwidth}
			\centering
			\captionof{table}[Example for the distinction between concept keys and node keys.]{Example for the distinction between concept and node keys.} 
			\begin{minipage}{0.5\textwidth}
				\centering
				\begin{tabular}{l|l|l}
					\textbf{concept} & \textbf{node key} & \textbf{concept key}\\
					\hline
					pain pattern				& [0]					& \underline{[0]} \\ 
					cardinal symptom			& [0,0] 				& \underline{[0,0]} \\
					radiating pain				& [0,1] 				& \underline{[0,1]} \\
					localization				& [0,0,0], [0,1,0] 		& \underline{[0,x,0]} \\
					intensity				& [0,1,1]		 		& \underline{[0,1,1]} \\
					spine						& [0,0,0,0], [0,1,0,0]	& \underline{[0,x,0,0]} \\
					head						& [0,0,0,1], [0,1,0,1]	& \underline{[0,x,0,1]} \\
					shoulder/arm/hand			& [0,0,0,2], [0,1,0,2]	& \underline{[0,x,0,2]} \\
					high						& [0,1,1,0]		 		& \underline{[0,1,1,0]} \\
					medium						& [0,1,1,1]		 		& \underline{[0,1,1,1]} \\
				\end{tabular}
			\end{minipage}
		\end{minipage}
	\end{scriptsize}
\end{center}
For illustration purposes, the corresponding graph with node and concept keys (underlined) is shown in Figure \ref{Abb:BegriffKnotenschluessel}.

\begin{figure}[!h]
	\centering
	\includegraphics[width=1.00\textwidth]{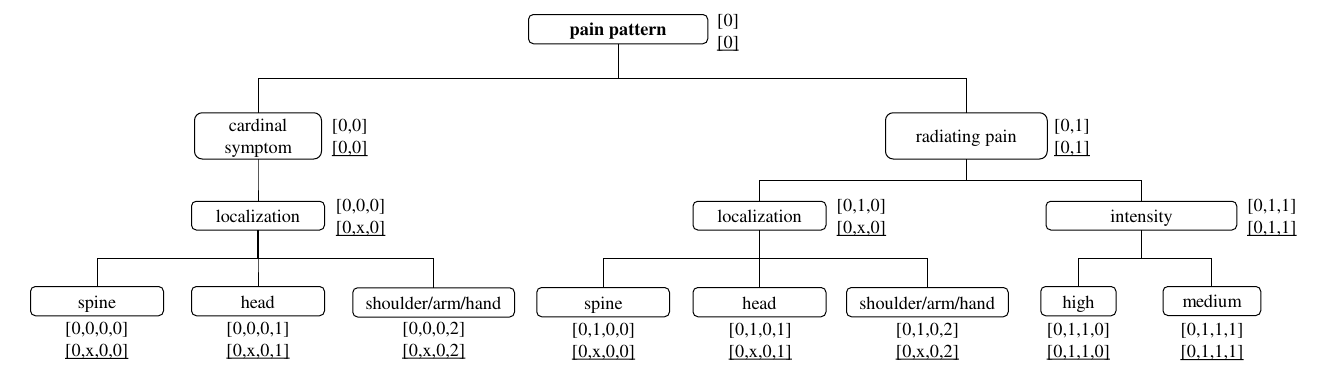}
	\caption[Example graph for the distinction between concept keys and node keys.]{Example graph for the distinction between concept keys and node keys.}
	\label{Abb:BegriffKnotenschluessel}
\end{figure}

\end{Bsp}

Thus, every more specific key can be unified with a more general key, because the more general key has to be initial key of the more specific key.
With the processing using partial unification, all necessary parts of the hierarchy can be reconstructed unambiguously.
\par
The keys are here represented as terms in list syntax.
Other syntactical representations are of course possible.
The list unification of the terms can be implemented very effectively.
Because independent variables and constants are used, it is merely a partial matching.

\subsection{Example \enquote{Simple anamnesis}}
This chapter presents an example for the specifics of the knowledge representation and the indexing algorithm.
The formal details will be explained in the following chapters.
\par
\begin{figure}[!h]
	\centering
	\includegraphics[width=0.90\textwidth]{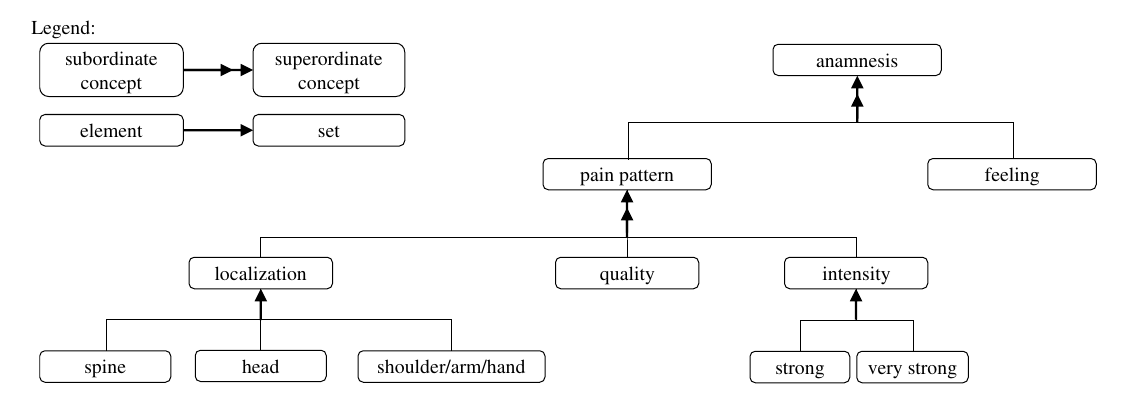}
	\caption[Simple anamnesis tree as well as uniform knowledge representation.]{Simple anamnesis tree as well as uniform knowledge representation.}
	\label{Abb:Anamnesebaum}
\end{figure}

A highly simplified anamnesis is examined.
The terminology is shown in Figure \ref{Abb:Anamnesebaum}.
Inheritance takes place from the roots to the leaves.
For example, the node \enquote{strong} indicates that a strong pain intensity was detected during the anamnesis.

Now the indexing algorithm has to assign keys to all concepts.
The details of this process are described in the algorithm specification (Chapter \ref{Abschnitt:BeschreibungAlg}).
After all concepts have been assigned keys,
the hierarchy in Figure \ref{Abb:Anamnesebaum} can be represented with the keys shown in Listing \ref{listingBaum}.

\lstset{numbers=left,captionpos=b}
\begin{lstlisting}[float,floatplacement=H,frame=single,numbers=none,caption={Possible representation of the tree with keys.},label=listingBaum]
([0] "anamnesis" ([0,0] [0,1]))
([0,0] "pain pattern" ([0,0,1] [0,0,2] [0,0,3]))
([0,1] "feeling")

([0,0,1] "localization" ([0,0,1,0] [0,0,1,1] [0,0,1,2]))
([0,0,1,0] "spine")
([0,0,1,1] "head")
([0,0,1,2] "shoulder/arm/hand")

([0,0,2] "quality")

([0,0,3] "intensity" ([0,0,3,0] [0,0,3,1]))
([0,0,3,0] "strong")
([0,0,3,1] "very strong")
\end{lstlisting}
Usually the keys also contain variables. 
Nodes in the hierarchy that are not leaves are also given additional references to the concepts underneath. 
This hierarchy is simple insofar as each concept only appears once. 
In general this is not the case. 
\par
To store an aspect of the anamnesis, a sequence of child nodes is selected, starting from the root node with the key $[0]$ and ending when the most specific valid node is reached. 
This node has a key that represents the chosen node and the whole path from the root to the node itself.
\par
Now the obtained key, e.g. $[0,0,1,1]$, can be stored in the database with a unique primary key id.
From this key it is possible to unambiguously reconstruct every valid key of the hierarchy.
In this case the key $[0,0,1,1]$ specifies the concept \enquote{head}.
The key $[0,0,1]$ is unifiable (Definition \ref{Def_unifizierbar}) with $[0,0,1,1]$.
It stands for \enquote{localization}.
From here one arrives at \enquote{pain pattern} and finally \enquote{anamnesis}.
A node is specified through the path to itself (in the example the concept itself, without the path, is sufficient for identification but it should be noted that in general, concepts can appear multiple times).

\subsection{Uniform representation}
The hierarchy in Figure \ref{Abb:Anamnesebaum} shows two structures: an inheritance structure (characterized by two-headed arrows) and an element-set-relationship (characterized by single-headed arrows).
With the semantic indexing, both structures can be represented and treated uniformly.
\par
The uniform representation enables a consistent syntax of representation over multiple implementations.
It formulates syntactical instructions for describing the representation and thus is an integral property of a knowledge database. 
The representation allows for concept and node keys with the contained inheritance paths to be used independently from specific inference methods or implementations.
\subsection{Example \enquote{concept key}, \enquote{partial unification of subtrees}}
\label{subsection:BspRepBegriffe}
\begin{figure}[!h]
	\centering
	\includegraphics[width=1.0\textwidth]{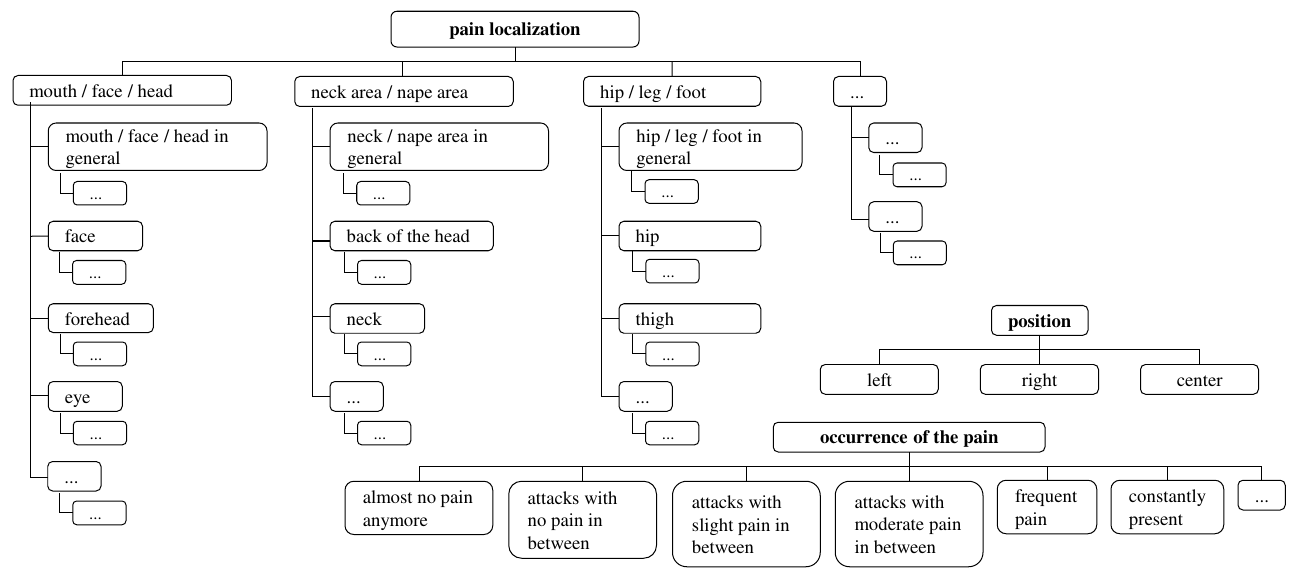}
	\caption[Excerpt from the domain's concept hierarchies for \enquote{pain localization}, \enquote{occurrence of the pain} and \enquote{position}]{Excerpt from the domain's concept hierarchies for \enquote{pain localization}, \enquote{occurrence of the pain} and \enquote{position}.}
	\label{Abb:Begriffsterminologie}
\end{figure}

Figure \ref{Abb:Begriffsterminologie} shows the concept hierarchy of a domain. The nodes in the example are important concepts that are ordered according to the domain knowledge.

The concept hierarchy \enquote{position} enables a more precise specification of the concepts in \enquote{pain localization}.
This is possible through the partial unification.
The concept hierarchy \enquote{occurrence of the pain} also allows a more precise specification of the whole pain profile (Chapter \ref{subsct:multiaxial}).
\par
An excerpt from the representation of the example from Figure \ref{Abb:Begriffsterminologie} is shown in Listing \ref{listing1}\footnote{The prefixed letter $a$ allows for an easier identification of the anamnesis subtree. In later examples, $d$ for diagnosis and $t$ for therapy are also used.}.

\noindent
\begin{minipage}{\linewidth}
\lstset{numbers=left,captionpos=b}
\begin{lstlisting}[frame=single,numbers=none,caption={Syntactical representation of the excerpt from the concept hierarchies for \enquote{pain localization}, \enquote{occurrence of the pain} and \enquote{position}.},label=listing1]
([a,x,0,1] "pain localization" ([a,x,0,x,0] [a,x,0,x,1] [a,x,0,x,6]))
	([a,x,0,x,0] "mouth / face / head" (a,1,0,1,0,0] [a,1,0,1,0,1] 
	 [a,1,0,1,0,2] [a,1,0,1,0,3] [a,1,0,1,0,4]))
		([a,1,0,1,0,0] "mouth / face / head in general" ([a,1,0,1,x,x,1,1]  
		 [a,1,0,1,x,x,1,2] [a,1,0,1,x,x,1,3]))
		([a,1,0,1,0,1] "face" ([a,1,0,1,x,x,1,1] [a,1,0,1,x,x,1,2] 
		 [a,1,0,1,x,x,1,3])) 
		([a,1,0,1,0,2] "forehead" ([a,1,0,1,x,x,1,1] [a,1,0,1,x,x,1,2] 
		 [a,1,0,1,x,x,1,3])) 
		([a,1,0,1,0,3] "eye" ([a,1,0,1,x,x,1,1] [a,1,0,1,x,x,1,2])) 
		(...)
	(...)

([a,1,0,1,x,x,1,1] "left")
([a,1,0,1,x,x,1,2] "right")
([a,1,0,1,x,x,1,3] "middle")
	
([a,x,0,8] "occurrence of the pain" ([a,x,0,8,0] [a,x,0,8,1] [a,x,0,8,2] 
 [a,x,0,8,3] ...))
	([a,x,0,8,0] "almost no pain anymore")
	([a,x,0,8,1] "attacks with no pain in between")
	([a,x,0,8,2] "attacks with slight pain in between")
	(...)
\end{lstlisting}
\end{minipage}
 
\section{Indexing algorithm}\label{sct:MainIndexingAlgorithm}
After the presentation of the theoretical groundwork in the previous chapter, the indexing algorithm will now be presented\footnote{Algorithmic presentation and proofs follow an unpublished report by U. Petersohn and J. Lehmann of Technische Universität Dresden, faculty of computer science, 2005.}.
\subsection{Correctness and completeness of an indexing algorithm}
\label{sct:KorrektheitVollstaendigkeit}
Prior to the description of the algorithm, the properties of correctness and completeness will be defined.
\begin{Def}[Correctness]
	\label{Korrektheit}
	An indexing algorithm is correct if it fulfills the following properties:
	\begin{enumerate}
		\item Every concept is assigned a key such that for any two keys $\key_{1}$ for concept $B_{1}$ and $\key_{2}$ for concept $B_{2}$: $inst({\underline{\key}_{B}}_{1}) \cap inst({\underline{\key}_{B}}_{2}) = \emptyset$.
		\item Let $\underline{\key}_{B}$ be a key of length $m$ for an arbitrary concept $B$. Every node $K$ with the concept $B$, that is reached in the concept hierarchy via the path $K_{1} \rightarrow ... \rightarrow K_{n} \rightarrow K$, is assigned a node key $\key_{K}$ that fulfills the following conditions:
		\begin{enumerate}
			\item $\key_{K}$ is an instance of $\underline{\key}_{B}$.
			\item Either $K$ is the root of the hierarchy or there exists an $i$ with $1\leq i \leq m$ such that $K_{n}$, the parent node of $K$, has the node key $\key_{i}$, i.e: $K_{n}$ is an initial key of $K$.
			\item All $\key\left[0, j\right]$ with $1 \leq j \leq m$ are not keys of nodes that are not in the set $\left\{K_{1}, ..., K_{n}, K\right\}$.
		\end{enumerate}	
	\end{enumerate}
\end{Def}

The first criterion in the definition enables the clear distinction between different concept keys.
The second criterion deals with node keys.
2a connects node keys and concept keys.
2b and 2c indicate that for a node $K$ all initial keys of $\key_{K}$ appear directly on the path to $K$ and do not appear on any other path.
\par
Additional properties follow from the definition of correctness. They will be proven below.

\begin{Satz}[Correctness]
	\label{SatzKorrektheit}
	If an indexing algorithm is correct according to Definition \ref{Korrektheit}, it also has the following properties:
	\begin{enumerate}
		\item For two nodes $K$ and $K'$ with $K \neq K'$ it is $inst(\key_{K}) \cap inst(\key_{K'})=\emptyset$.
		\item For two nodes $K$ and $K'$, with $K$ parent node of $K'$, $\key_{K'}$ is partially unifiable with $\key_{K}$.
		\item For two concepts $B$ and $B'$, and a node of concept $B$ that is the parent node of a node of $B'$, $\underline{\key}_{B'}$ is partially unifiable with $\underline{\key}_{B}$.
	\end{enumerate}
\end{Satz}

The items of Proposition \ref{SatzKorrektheit} will be proven consecutively (numbering as in Proposition \ref{SatzKorrektheit}).

\begin{enumerate}
	\item Proof by contradiction: Let there be two nodes $K$ and $K'$ with $K \neq K'$ and a key $\key$, that is an instance of both $\key_{K}$ and $\key_{K'}$. The proof is split in two parts:
	\begin{enumerate}
		\item $K'$ is on the path from $K$ to the root: \newline
		Then, according to Definition \ref{Korrektheit} 2b, $\key_{K}$ is an initial key of $\key_{K'}$, and therefore $\key_{K} \neq \key_{K'}$.
		\item $K'$ is not on the path from $K$ to the root: \newline
		Then, according to Definition \ref{Korrektheit} 2c (for the special case $j=m$), $\key_{K}$ is no node key of a node outside of the path $K_{1} \rightarrow ... 				\rightarrow K_{n} \rightarrow K$. Because $K'$ is outside of that path, $\key_{K} \neq \key_{K'}$ follows.
	\end{enumerate}
	\item According to Definition \ref{Korrektheit} 2b, $\key_{K}$ is an initial key of $\key_{K'}$. The claim follows directly.
	\item Let $K$ be the node with concept $B$ and $K'$ the node with concept $B'$. Following from the preconditions, $K$ is a parent node of $K'$. According to Definition \ref{Korrektheit} 2b, $K$ is initial key of $K'$, therefore $\key_{K'}$ is partially unifiable with $\key_{K}$. Furthermore, by Definition \ref{Korrektheit} 2a, $\key_{K}$ is an instance of $\underline{\key}_{B}$ and $\key_{K'}$ is an instance of $\underline{\key}_{B'}$. Therefore $\underline{\key}_{B'}$ is partially unifiable with $\underline{\key}_{B}$.
\end{enumerate}

The criterion for completeness is less complex than correctness, as shown by the following definition.

\begin{Def}[Completeness] 	
	An indexing algorithm is complete if it creates an index for each given concept hierarchy.
\end{Def}

\subsection{Description of the indexing algorithm}
\label{Abschnitt:BeschreibungAlg}
The algorithm uses the two operations generalization and expansion of keys.
\begin{Def}[Generalization of keys]
	Let $\key_{1}, ..., \key_{m}$ be a finite sequence of keys.
	The generalized key $\gkey$ of these keys is defined as follows:
	\\
	$\gkey\left[i\right] = 
	\begin{cases}
	n, & \text{if for all keys } \key_{j} \text{ with length} \geq i  \text{: }  \key_{j}\left[i\right]=n\ (n \in \mathbb{N}) \\
	x, & \text{if a key } \key_{j} \text{ with length} \geq i \text{ exists and the above condition does not apply} \\ 
	\end{cases}$
	\newline
	$\gkey$ has the same length as the longest key $\key_{j}\left(1 \leq j \leq m\right)$.
	\label{AdditionSchluessel}
\end{Def}

\begin{Bsp}[Generalization of keys]
	This example shows the generalization of the keys $\left[0,0,2,x,8\right]$ and $\left[0,x,8,x,8\right]$. \\\\
	$\begin{array}{ccccc}
	0	&	0	& 	2 	&	x 	&	8 \\
	0	&	x	&	8	&	x	&	8 \\
	\hline
	0	& x	&	x	&	x	&	8
	\end{array}$
\end{Bsp}

The generalization is a consecutive comparison of all keys at a specific position. 
If there are different values at the same position or the variable $x$ appears, the result at that position is $x$.
If all values at a position are the same natural number, that number is kept.

\begin{Def}[Expansion of keys]	
	The expansion ${\key}'_{1}$ of a key $\key_{1}$ of length $m$ towards a key $\key_{2}$ of length $n$ with $m \leq n$ is defined as follows:
	\\
	${\key}'_{1}\left[i\right] = 
	\begin{cases}
	\key_{1}\left[i\right], & \text{if } i\leq m \\
	\key_{2}\left[i\right], & \text{if } m < i \leq n
	\end{cases}$ \\
	${\key}'_{1}$ has the length $n$.
\label{Def:ErweiternSchluessel}
\end{Def}

During the expansion of a key $\key_{1}$ towards another key $\key_{2}$, $\key_{1}$ is filled up with values from $\key_{2}$ so that both have the same length.

\begin{Bsp}[Generalization and final expansion of keys]
	After the generalization of a set of keys and the subsequent expansion of the shorter keys towards the generalized key, it is clear that all created keys are instances of the generalized key.
	Combining the previous examples for generalization and expansion results in the following:  \\\\
	$\begin{array}{ccccc}
	0	&	0	& 2 &	x &	\fbox{8} \\
	0	&	x	&	8	&	x	&	8 \\
	\hline
	0	& x	&	x	&	x	&	8  
	\end{array}$ \\
	
	The boxed $8$ was added during the expansion towards the generalized key. 
	The keys $\left[0, 0, 2, x, 8\right]$ and $\left[0, x, 8, x, 8\right]$ are instances of $\left[0, x, x, x, 8\right]$.
\end{Bsp}

As an abbreviation, a notation for the \enquote{parent nodes of a concept} is introduced (Definition \ref{def:parent-node-set}).
The quotation marks are used because a concept itself is not a node, and therefore also does not have a parent node.
\begin{Def}[\enquote{Parent nodes of a concept}]
	For a concept $B$, $parents(B)$ is defined as the set of all nodes in the concept hierarchy that have a child node with concept $B$.
	\label{def:parent-node-set}
\end{Def}

\begin{Bsp}[\enquote{Parent nodes of a concept}]
	In Figure \ref{Abb:BegriffKnotenschluessel}, $parents$(\enquote{localization}) is a set with the two nodes \enquote{cardinal symptom} and \enquote{radiating pain}.
\end{Bsp}

Based on the previous definitions, the algorithm can now be specified.
Input for the algorithm is the given concept hierarchy.
Additionally, for each node a number is stored and initialized with $0$.
Furthermore, every node in the concept hierarchy stores the respective node key.
\begin{Alg}[Indexing algorithm]
	The indexing algorithm can be specified as follows.
	\begin{tabbing}
		\hspace*{3.0cm}\=\kill
		Input: 						    \>concept hierarchy $H$ with a counter $0$ in each node \\
		Output: 						\>all concept keys and node keys \\
		Initialization: 				\>the root node of $H$ and its concept get the key $\left[0\right]$\\
	\end{tabbing}
	The following operations are repeated until every concept has a key:
	\begin{itemize}
		\item OP I - selection operation: \\ 
		Choose a concept $B$ such that all elements of $parents(B)$ already have a node key.
		\item OP II - derivation operation: \\
		For each element $K \in parents(B)$ a key is generated by appending the counter of $K$ to the key $\key_{K}$.
		Next, the counter of $K$ is raised by 1.
		If a node appears twice or more with the same path length, two or more keys are created.
		Those are generalized in OP III.
		An expansion (OP IV) is not necessary because both keys have the same length.
		The result is a concept key.
		The node keys are instances of that key and are updated in the concept hierarchy.
		\item OP III - generalization operation: \\
		All generated keys are generalized according to Definition \ref{AdditionSchluessel}.
		If the generalized key has instances in common with another concept key that was already created, more numbers are appended until this is no longer the case.
		The resulting key is $\key_{B}$.
		\item OP IV - expansion operation: \\
		Now all node keys are expanded towards $\key_{B}$.
		If keys were created and the generalization resulted in concept keys of differing length, those are expanded according to Definition \ref{Def:ErweiternSchluessel}.
		In this manner, all node keys are obtained.
		\item OP V - addition operation: \\
		The obtained node keys are added to the respective nodes in $H$.
	\end{itemize} 
	\label{Indizierungsalgorithmus}
\end{Alg}

The described indexing algorithm is illustrated with a detailed example in Chapter \ref{subsct:BspIndizierung}.\par\medskip
\subsection{Proof of the correctness of the algorithm}
\label{sct:proof-correctness}
\begin{Satz}[Correctness]
	Algorithm \ref{Indizierungsalgorithmus} is correct.
\end{Satz}

\begin{figure}[!h]
	\centering
    \includegraphics[width=0.2\textwidth]{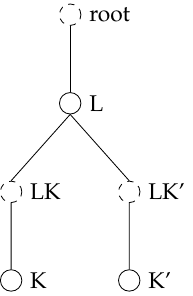}
	\caption[Illustration for the proof of correctness.]{Illustration for the proof of correctness.}
	\label{Abb:Korrektheitsbeweis}
\end{figure}

To prove the correctness of the algorithm, all properties from Definition \ref{Korrektheit} have to be shown.
The following chapter uses the same notations and numberings as the definition.
\begin{enumerate}
	\item This property follows directly from OP III and IV. The algorithm demands that a created concept key must not have instances in common with any other key.
	\item 
	\begin{enumerate}
		\item $\underline{\key}_{B}$  is the generalization of all node keys with the concept $B$. Therefore, every node key with $B$ is an instance of $\underline{\key}_{B}$.
		\item For the parent node $K_{n}$ of a node $K$, the node key $\key_{K}$ is generated in the algorithm by appending a number to $\key_{{K}_{n}}$. Then the key is expanded. Clearly, $\key_{{K}_{n}}$ is an initial key of $\key_{K}$.
		\item As a proof by contradiction it is assumed that there exists a node $K'$ whose key $\key_{K'}$ is an initial key of  $\key_{K}$, but which is not part of the path from $K$ to the root.
		Then there is a node $L$ in the concept hierarchy at which the paths from $K$ and $K'$ to the root intersect.
		Let $\key_{L}$ have the length $n$.
		$L$ has a child node  $LK$ on the path from $K$ to the root and a child node $LK'$ on the path from $K'$ to the root.
		Figure \ref{Abb:Korrektheitsbeweis} illustrates this.
		In the figure, dashed lines symbolize potentially multiple nodes.
		Because every node of the concept hierarchy by definition only has nodes with different concepts as children, $LK$ and $LK'$ belong to different concepts.
		Therefore, the counters that were appended during the computation of $\key_{LK}$ and $\key_{LK'}$ in OP II are different.
		Hence, $\key_{LK}$ and $\key_{LK'}$ differ at position $n+1$ of the keys.
		By applying 2b multiple times for $K$ and $K'$ it follows that $\key_{LK}$ is an initial key of $\key_{K}$ and $\key_{LK'}$ is an initial key of  $\key_{K'}$.
		Thus, $\key_{K}$ and $\key_{K'}$ also differ at position $n+1$ and $\key_{K'}$ can not be an initial key of $\key_{K}$.
	\end{enumerate}
\end{enumerate}

\subsection{Proof of the completeness of the algorithm}
\label{subsct:Proofcompleteness}
\begin{Satz}[Completeness]
	Algorithm \ref{Indizierungsalgorithmus} is complete.
\label{SatzVollstaendigkeit}
\end{Satz}
For the proof of completeness it is necessary to show that the algorithm computes an output in finite time for each input.
The operations II to V are simple computations that clearly are always executable in finite time.
Therefore, proving operation I is sufficient.
It states that a concept $B$ has to be found such that all elements of $parents(B)$ already have node keys.
To prove completeness it has to be shown that for all cases such a concept $B$ exists.
\par
For the proof it is assumed that no such $B$ can be chosen.
Then it is shown that in that case the dependency graph of the concept hierarchy is not free of cycles.
This violates the definition of concept hierarchies (Definition \ref{Begriffshierarchie}).
\par
\begin{figure}[!h]
	\centering
	\includegraphics[width=0.5\textwidth]{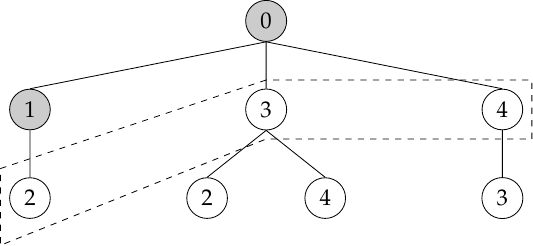}
	\caption[Illustration for the proof of completeness.]{Illustration for the proof of completeness.}
\label{Abb:Vollstaendigkeitsbeweis}
\end{figure}

The indexing algorithm computes the node keys based on operation I such that every newly indexed node already has an indexed parent.
It is assumed that the algorithm reaches a point at which there is no concept $B$ for which all elements of $parents(B)$ are already indexed.
In that step there is a set of nodes $K_{1}, ..., K_{n}$ whose parent nodes are already indexed. 
Let the corresponding concepts be $B_{1}, ..., B_{n}$.
The subtrees spanned by  $K_{1}, ..., K_{n}$ will be called $T_{1}, ..., T_{n}$.
Figure  \ref{Abb:Vollstaendigkeitsbeweis}  illustrates the situation.

The gray nodes have already been indexed.
The nodes framed by the dashed line are $K_{1}, ..., K_{3}$.
\par
For all $B_{i}~(1\leq i\leq n)$ there are, as required, nodes in $parents(B_{i})$ that are not in the set $\left\{K_{1}, ..., K_{n}\right\}$ and also not part of the already indexed nodes.
\par
Let $L$ be such a node with concept $B_{1}$
This node appears in one of the subtrees spanned by $K_{1}, \dots, K_{n}$.
If $L$ appears in the subtree $T_{1}$ then there is a cycle in the dependency graph because $K_{1}$ is a node with concept $B_{1}$ and in the tree there is the node $L$ below $K_{1}$ which also has concept $B_{1}$.
In the dependency graph, an arrow would lead from $B_{1}$ (via eventual intermediate nodes) back to $B_1$.
Therefore $L$ can not be part of $T_{1}$.
\par
Then $L$ has to appear in at least one other subtree $T_{{i}_{1}}(i_{1}\neq 1)$.
Therefore, there is a path $B_{{i}_{1}} \rightarrow \cdots \rightarrow B_{1}(i_{1} \neq 1)$ in the dependency graph (because  $B_{1}$ is below $B_{{i}_{1}}$ in the concept hierarchy), so $B_{1}$ depends on $B_{{i}_{1}}$.
\par
Starting from $B_{{i}_{1}}$ the same argument can be applied again.
If there is a node with concept $B_{{i}_{1}}$ that appears in $T_{1}$  and $T_{{i}_{1}}$  then the dependency graph contains a cycle.
It follows that a subtree $T_{{i}_{2}}(i_{2} \neq i_{1}, i_{2} \neq 1)$  exists in which a node with concept $B_{{i}_{1}}$ appears.
Then the dependency graph contains a path $B_{{i}_{2}} \rightarrow \cdots \rightarrow B_{{i}_{1}} \rightarrow \cdots \rightarrow B_{1}(i_{2} \neq i_{1}, i_{2} \neq 1, i_{1} \neq 1)$.
\par
Applying this argument $n$ times results in a path $B_{{i}_{n}} \rightarrow \cdots \rightarrow B_{{i}_{n-1}} \rightarrow \cdots ... \cdots \rightarrow B_{{i}_{1}} \rightarrow \cdots \rightarrow B_{1}(i_{j} \neq i_{k}, i_{j} \neq 1 \text{~for all~} j, k \text{~mit~} 1\leq j, k \leq n, j\neq k)$ in the concept hierarchy. But because there are only $n$ different concepts, one of the concepts has to appear multiple times in that path.
Therefore there is a cycle in the dependency graph.
This violates the definition of concept hierarchies.
\par
The proofs of correctness and completeness show that the algorithm always produces the correct result.
\subsection{Example for the operating principle of the indexing algorithm}
\label{subsct:BspIndizierung}
\subsubsection{Example}
The previous remarks will now be demonstrated with an example for the operating principle of the algorithm.
\paragraph{Initialization}
Starting point is the concept hierarchy that is initialized by the algorithm.
The root node and its concept can already be indexed as shown in the following figure.
The counters stored for each node are initialized with $0$.
\begin{figure}[H]
	\centering
	\includegraphics[width=13.5cm]{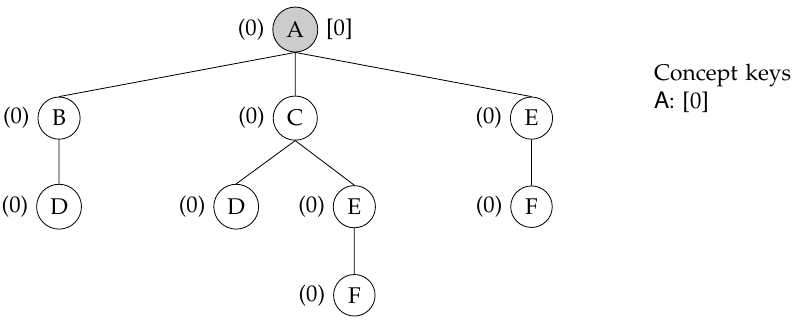}
	\caption[Initialization concept $A$.]{Initialization concept $A$.}
\end{figure}

\paragraph{Step 1}
In this step the concept $B$ is chosen (in accordance with $OP\text{ }I$).
$parents(B)$ contains only one node, the root node.
The counter of the root node is appended to its key, resulting in $\left[0, 0\right]$.
Then the counter of the root node is raised ($OP\text{ }II$).
Because there is only the key $\left[0, 0\right]$, $OP\text{ }III$ (generalization) and $OP\text{ }IV$ (expansion) of the algorithm are trivial. 
The resulting concept key for $B$ is $\left[0, 0\right]$.
In $OP\text{ }V$ the node key $\left[0, 0\right]$ is updated in the concept hierarchy.
\begin{figure}[H]
	\centering
	\includegraphics[width=13.5cm]{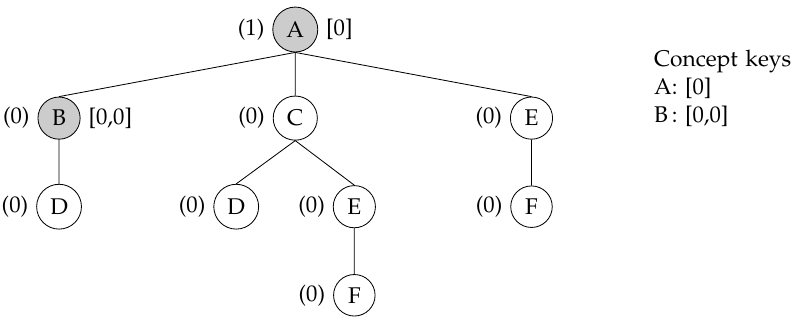}
	\caption[Step 1 (concept $B$).]{Step 1 (concept $B$).}
\end{figure}

\paragraph{Step 2}
Now concept $C$ is chosen ($OP\text{ }I$).
Analogous to the previous step, by appending the counter of the root node to its key, the key $\left[0, 1\right]$ is obtained.
Like in the previous step, $OP\text{ }III$ and $OP\text{ }IV$ are trivial.

\begin{figure}[H]
	\centering
	\includegraphics[width=13.5cm]{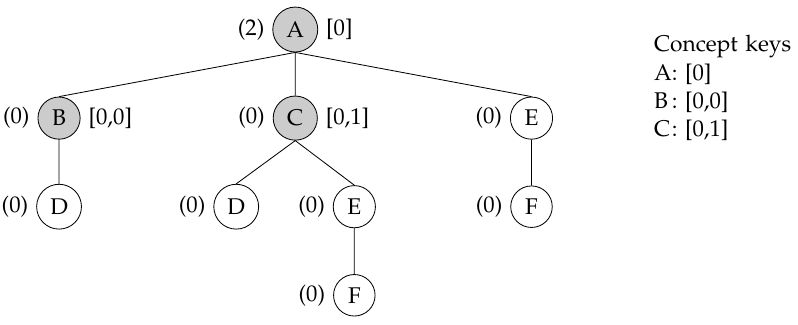}
	\caption[Step 2 (concept $C$).]{Step 2 (concept $C$).}
\end{figure}

\paragraph{Step 3}
Now concept $D$ is chosen.
Because $D$ appears twice, two keys are created: $\left[0, 0, 0\right]$ and $\left[0, 1, 0\right]$.
These are generalized in $OP\text{ }III$ (see auxiliary calculation).
An expansion ($OP\text{ }IV$) is not necessary because both keys have the same length.
The resulting concept key is $\left[0, x, 0\right]$.
The node keys are instances of this key and are updated in the concept hierarchy.

\begin{figure}[H]
	\centering
	\includegraphics[width=14.25cm]{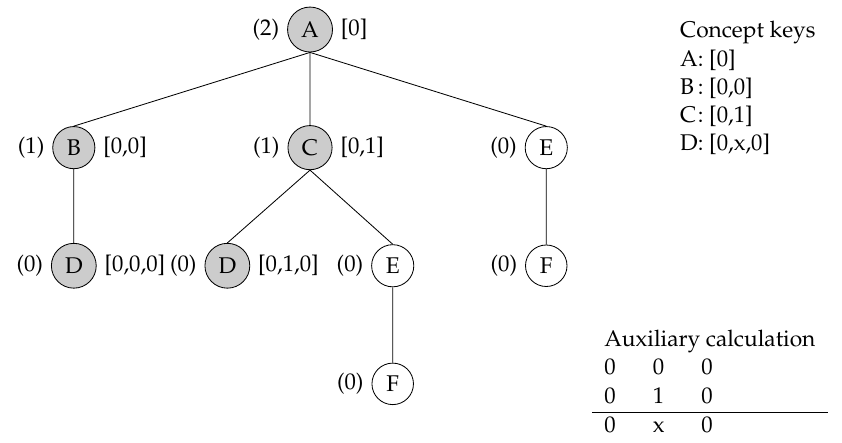}
	\caption[Step 3 (concept $D$).]{Step 3 (concept $D$).}
\end{figure}

\paragraph{Step 4}
In this step $E$ is the only concept that can be chosen.
Analogous to the previous steps the keys $\left[0, 1, 1\right]$ and $\left[0, 2\right]$ are generated.
The generalization produces the concept key $\left[0, x, 1\right]$.
The two created keys do not have the same length, so in accordance with $OP\text{ }IV$ the keys are expanded towards the generalized key: The boxed $1$ is added to $\left[0, 2\right]$.
This results in the two node keys $\left[0, 2, 1\right]$ and $\left[0, 1, 1\right]$.

\begin{figure}[H]
	\centering
	\includegraphics[width=14.75cm]{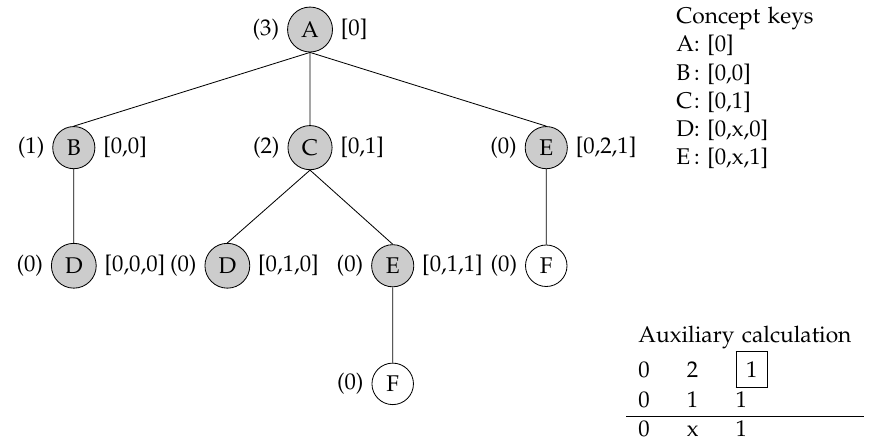}
	\caption[Step 4 (concept $E$).]{Step 4 (concept $E$).}
\end{figure}

\textbf{Step 5} \newline
The last concept $F$ is also indexed as described.
At this point it should be noted that in $OP\text{ }III$ it has to be checked whether the obtained key shares instances with other concept keys.
If this is the case, more numbers must be appended to the generalized key, so that there are no shared instances.
Because all concepts are fully indexed, the indexing algorithm now terminates.

\begin{figure}[H]
	\centering
	\includegraphics[width=14.75cm]{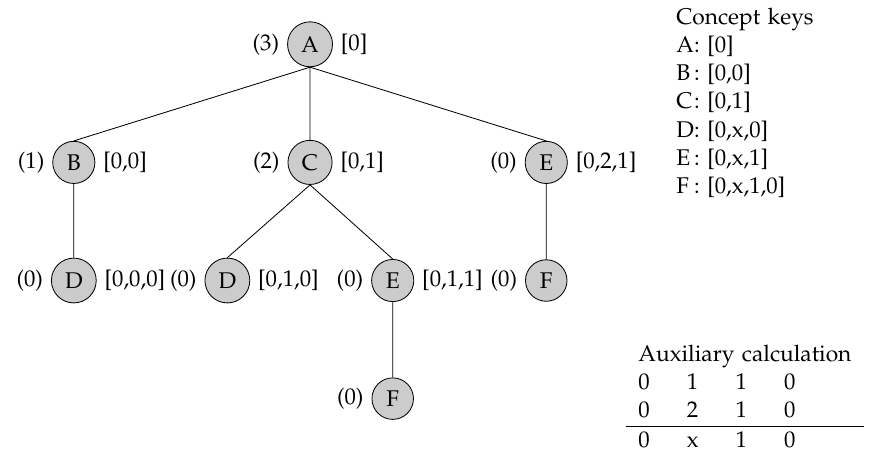}
	\caption[Step 5 (concept $F$).]{Step 5 (concept $F$).}
\end{figure}

\subsubsection{Remarks}

\begin{itemize}
	\item The algorithm offers an indexing. However, there exist other indexing solutions, because in each path nodes can appear in different orders. It is not practical to consider all possible indexings. If necessary, the input concept hierarchy can be suitably sorted beforehand.
	\item If needed, side conditions for the index creation can be built in, as long as correctness and completeness are not violated. Biunique renaming is possible.
	\item If concepts are distributed across multiple levels of a concept hierarchy, this can possibly lead to a large amount of variables that have to be inserted. Ideally each concept should be spread across only one level. This provides a clearer structure. More complex compositions of concepts can be implemented more clearly with a multiaxial description (Chapter \ref{subsct:multiaxial}).
\end{itemize}

\section{Multiaxial modeling of concept hierarchies}
\label{subsct:multiaxial}

\subsection{Forced hierarchization versus independent trees}
Prerequisite for the semantic indexing is the existence of a concept hierarchy, or the possibility of modeling a concept hierarchy (Chapter \ref{subsectionGrundbegriffe}).
One possible problem of semantic indexing is a disadvantageous forced hierarchization of aspects in the knowledge acquisition and formalization, or rather, that many variables $x$ have to be introduced to represent concepts used on multiple levels.
This happens in particular when the structuring of knowledge bases in the modeling phase is unclear and can be problematic.
The recommended solution is a multiaxial description and representation.
\par
Frequently a concept hierarchy can only model one specific aspect of reality.
Still, a powerful system should be able to connect multiple aspects (axes).
This connection is declared by the multiaxial description which can be easily handled with semantic indexing.
For each individual axis there exists a concept hierarchy that can be indexed.
Each concept hierarchy has a unique name and key.
The axes are then combined conjunctively (Chapter \ref{subsct:Begriffshierarchien}).

\subsection{Concept hierarchies, uniaxial and multiaxial systems}
Terminologies and taxonomies have to represent the complete range of formulations and synonyms of the application domain.
\par
They serve the creation of concept hierarchies, the classification as well as the usage of adequate, semantically correct concepts on the basis of propositions of the domain.
\begin{Def}[System of order]
	Systems of order for concepts systematically map propositions to concept units.
	An order is formed through conceptual, systematic and semantic axes.
	Concept hierarchies can be defined.
\end{Def}

In general the following requirements have to be met by systems of order:

\begin{itemize}
	\item \textbf{Completeness:} \\
	The considered domain has to be represented completely, i.e. there must not be any missing concepts and it has to be possible to add new concepts.
	\item \textbf{Disjointness:} \\
	All concepts should be represented uniquely without overlaps.
	Redundancies should be avoided.
	If multiple identifiers are necessary, synonymic links and preferred identifiers can be added.
	Uniqueness of concepts has to be preserved.
	Homonyms and ambiguities should be avoided.
	\item \textbf{Classification:} \\
	The system of order has to be built to be consistent, free of contradictions and transparent, following a classification that is scientifically or practically recognized.
	If one alone is not enough, multiple classifications have to be allowed.
\end{itemize} 

Depending on the number of semantic axes, uniaxial and multiaxial systems are distinguished.

\begin{Def}[Monoaxial or uniaxial system of order]
	All concepts of interest are described with one axis.

	The domain of discourse is ordered by continuously adding one distinguishing feature per hierarchical level from the general to the specific.
	In general, the classes can not be combined with each other \cite{ZGIL05}.
	\label{Def:uniaxial}
\end{Def}

\begin{Def}[Multiaxial system of order]
	To systematize concepts and classes, multiple axes are used.
	A multiaxial system of order is based on a category structure.
	Concepts of multiple categories or semantic axes are combined to express one complex concept.
	Every semantic axis corresponds to another area of information \cite{ZGIL05}.
	\label{Def:multiaxial}
\end{Def}

Following Definition \ref{Def:allgspezieller}, the concept hierarchies (Definition \ref{Begriffshierarchie}) are ordered  with the relations \enquote{more specific} and \enquote{more general}.
These concept hierarchies are also the input for the semantic indexing algorithm (Chapter \ref{Abschnitt:BeschreibungAlg}).

\subsection{Extension with multiaxial descriptions}
\label{subsub:multiaxial}
\begin{figure}[!h]
	\centering
	\includegraphics[width=0.90\textwidth]{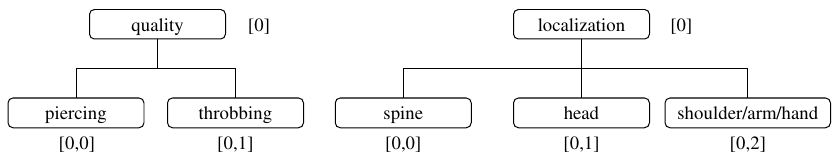}
	\caption[Conjunctive multiaxial description with two axes]{Conjunctive multiaxial description with two axes.}
	\label{multiaxialeBeschreibung}
\end{figure}

In the example in Figure \ref{multiaxialeBeschreibung}, the not yet mentioned \enquote{pain quality} is an aspect of the pain that can be surveyed independently from the localization.
If, for example, it is known that the patient feels a strong piercing pain, the knowledge base does not yet have to know if it is located at the temples and vice versa.
If this aspect is modeled in the tree in Figure \ref{multiaxialeBeschreibung}, the subtree \enquote{pain quality} can be either recorded above the localization, so that each of these quality nodes has a subtree with the complete localization tree (not just each leaf node, because the detailing can be stopped early), or the other way around.
That would make the tree less clear and harder to manage.
This can be solved through defined levels, i.e. levels 1 to 5 specify the localization and levels 6 and 7 the quality.
That makes it necessary to introduce additional variables $x$.
Furthermore, each aspect would need a defined depth, even when, for example at one position, certain aspects do not necessarily have to be described in such detail.
This means that through a possible \enquote{forced hierarchization}, a knowledge base could get unnecessarily complex and make derivations harder.
In that case it is better to represent different aspects in independent trees and link them to each other.
\par
For each individual axis there exists a concept hierarchy that can be indexed.
Every concept hierarchy can be given a unique name.
Figure \ref{multiaxialeBeschreibung} shows two highly simplified concept hierarchies for pain quality and localization.
Let now $Q$ be the name of the concept hierarchy for quality and $L$ the name for localization.

To describe, for example, a piercing headache, both concept hierarchies can be combined conjunctively.
For this, the appropriate node keys are stored for each axis.
For the example in Figure \ref{multiaxialeBeschreibung}, $[(Q[0,0]),(L[0,1])]$ describes a piercing headache.
The notation indicates that the axes $Q$ and $L$ are assigned the node keys $[0,0]$ and $[0,1]$ respectively.
This principle can be expanded by allowing several multiaxes.
For example, if a patient has a piercing headache and an additional throbbing pain in his arm, this can be denoted as  $[(Q[0,0]), (L[0,1])], [(Q[0,1]), (L[0,2])]$.
Of course, other notations following the same principle are possible here.
\par
More important than the specific notation is the fact that multiaxial descriptions can be easily realized with already indexed concept hierarchies.
Compared to a uniaxial model, this has the benefit that multiple axes can be modeled independently from each other.
Furthermore, a concept hierarchy that includes all axes would grow very fast and could no longer be managed efficiently.
An additional benefit is that various axes can be combined as needed.
To expand the above example, it would be possible to declare other axes (e.g. topography, development over time, etc.), so that the symptoms can be described more precisely.

\section{Hierarchies of deduced concepts and d-concept descriptions}
\label{sct:BegriffshierarchienKonz}
\subsection{Hierarchies of d-concepts}
Besides the domain-specific \enquote{atomic} concept hierarchies discussed until now, hierarchies of deduced concepts are also in use and have to be indexed. Together with the deduced concept descriptions, they serve the representation of generic knowledge and inference. To improve readability, deduced concepts will be abbreviated as {d-concepts} in the following chapters.

\begin{Def}[Hierarchy of d-concepts / d-concept hierarchy]
	A hierarchy of {d-concepts} is represented by a concept hierarchy (Definition \ref{Begriffshierarchie}).
	Nodes $\dconx \in \dconset$ correspond to {d-concepts}.
	Every node has a unique concept key (Definition \ref{DefKnotenBegriff}), a textual description and references (concept keys) to a set of subordinate concepts.
	If edges exist between {d-concepts}, the respective nodes have to be partially unifiable (Definition \ref{Def_unifizierbar}).
	The graph must not contain any cyclical definitions.
	Within the hierarchy of {d-concepts}, the relations between the individual {d-concepts} are \enquote{more general than} and \enquote{more specific than} (Definition \ref{Def:specific}).
	\label{Arbeitsdef:Konzeptterminologie} 
\end{Def}

\begin{Def}[Deduced concept / {d-concept}]
	Deduced concepts (d-concepts) $\dconx \in \dconset$ are generic objects that refer to a set of instances.
	The instances that fulfill a {d-concept} are called the {d-concept} extension.
	Necessary and sufficient conditions for instances are declared by the {d-concept} description.
	\label{Def:Konzept}
\end{Def}

First, inferences with d-concepts will be considered.
A given situation description $\underline{x} \in \bset$ consists of a tuple of node keys at a certain time.
\par
A d-concept $\dconx$ itself is a unary predicate over a base set $\bset$:
\begin{tabbing}
	\hspace*{1cm}\=\hspace{5.0cm}\=\kill
	\> $\underline{x} \in \bset$, $\dconx \in \dconset$, $\dconx(\underline{x})$:      \> $\underline{x}$ belongs to $\dconx$ \\
	\> $\underline{x} \in M$, $\dconx \in \dconset$, $\neg \dconx(\underline{x})$: \> $\underline{x}$ does not belong to $\dconx$ \\
\end{tabbing}
To describe relations in the domain of discourse, the d-concept knowledge base  contains heuristic knowledge of the domain (domain knowledge) as well as generic knowledge.
\par
Heuristic knowledge still contains the subjectivity of the creator and thus will not in all cases be sufficiently accepted.
Generic knowledge represents important general relations and is of objective nature.
\subsection{Inheritance of d-concepts}
\label{subsection:InheritanceConcepts}
For hierarchies of {d-concepts}, their inheritance is important.
A child {d-concept} $\dconc_{j}$ inherits from a parent {d-concept} $\dconc_{i}$ if it is a more specific version of $\dconc_{i}$.
For example, if the set of situation descriptions that are classified as $\dconc_{j}$  is a subset of the descriptions classified as $\dconc_{i}$, all instances that fulfill $\dconc_{j}$  also fulfill $\dconc_{i}$.
The relation from $\dconc_{i}$ to $\dconc_{j}$ is therefore called \enquote{more general than}.
\begin{Def}[\enquote{More general than}, \enquote{more specific than}]
	A d-concept $\dconc_{i}(\underline{x})$ is \enquote{more general than} a d-concept  $\dconc_{j}(\underline{x})$ if $\forall \underline{x} \in \bset: \left[\dconc_{j}(\underline{x}) \Rightarrow \dconc_{i}(\underline{x})\right]$, denoted as $\forall \underline{x} \in \bset: \left[\dconc_{i}(\underline{x}) \geq \dconc_{j}(\underline{x})\right]$.
	The inverse relation is called \enquote{more specific than}.
	\label{Def:specific}
\end{Def}
The relation \enquote{more general than} is a pseudo-order.
For example, it is transitive: if
$\dconc_{i}(\underline{x}) \geq \dconc_{j}(\underline{x})$ and $\dconc_{j}(\underline{x}) \geq \dconc_{k}(\underline{x})$ then $ \dconc_{i}(\underline{x}) \geq \dconc_{k}(\underline{x})$.
As long as every {d-concept} has a unique name, the relation is also a partial order.
It is, for example, antisymmetric: if  $\dconc_{i}(\underline{x}) \geq \dconc_{j}(\underline{x})$ and $\dconc_{i}(\underline{x}) \leq \dconc_{j}(\underline{x})$ then $\dconc_{i}(\underline{x}) = \dconc_{j}(\underline{x})$.

\subsection{Descriptions of d-concepts}
\label{subsection:Konzeptbeschreibungen}
To describe {d-concepts}, the respective concept keys of the domain are used to specify the defined {d-concept} as accurately as possible.
Thus, an association between the attribute characteristics of the declared {d-concepts} is established.
Cyclical definitions are not allowed.

\begin{Def}[Deduced concept description / d-concept description]
	Every d-concept is declared by its d-concept descriptions $\dconc_{desc}$.
	The d-concept descriptions contain the necessary and sufficient conditions for instances.
	\label{Def:Konzeptbeschreibung}
\end{Def}
\par
The d-concept description implies:
\begin{itemize}
	\item For \enquote{atomic} concepts, the validity of the concept keys has to be determined by a partial unification with the node keys from the knowledge base or by querying the agent in the application.
	\item For d-concepts, the validity of their concept keys has to be determined via inference regarding all necessary and sufficient conditions of the {d-concept} description.
	\item Hierarchies of {d-concepts} use the \enquote{more specific than} and \enquote{more general than} relations.
	This also applies to the {d-concept} descriptions, in which subordinate descriptions are declared by specializing the superordinate descriptions and by inheritance.
	\item The {d-concepts} are declared with concept keys based on domain knowledge.
	On the basis of inheritance relationships between the keys, the necessary and sufficient conditions with the assignment of the concept keys are done in a way where maximally general {d-concept} descriptions (Definition \ref{Def:max}) appear on every hierarchical level and the relations \enquote{more general than/more specific than} (Definition \ref{Def:specific}) are not violated.
\end{itemize}

\begin{Def}[Maximally general {d-concept} description]
	A {d-concept} description  $\dconc_{desc}$ is maximally general if it does not cover any negative instances (misclassification) and  $\dconc_{desc} \geq \dconcprime_{desc}$ for all other {d-concepts} $\dconcprime_{desc}$ that also do not cover any negative instances.
	\label{Def:max}
\end{Def}

\section{Discussion of inference methods}
\label{section:ProblemSolving}
With the uniform representation using node and concept keys, classic inference methods can be executed effectively (even with databases).
Complex knowledge queries are possible.
For these, generic and case-based knowledge can be used together.
The obtained solutions are stored as instances in a database.
\subsection{Logic-based inference}
\label{section:LogicBasedInference}
For the d-concept descriptions, e.g. clauses can be used.
The representational paradigm matches (without proof) an expanded monadic predicate logic.
Now, the known algorithms for logical inference can be used.
From the introduced node and concept keys, index tables with the contained inheritance paths can be built.
Combined with a partial matching, this can be translated into an efficient implementation of logic-based inference.

\subsection{Concept-based reasoning}
\label{section:CoBR}
In general, the decidability and complexity of solutions depend on the logical language.
Horn clauses allow a compact representation and are efficiently decidable.
However, compared to the standard Horn resolution, the following problems appear:
\begin{enumerate}
	\item The user does not want to query every possible {d-concept} (e.g. diagnosis). Instead, the system should automatically find all valid {d-concepts}.
	\item Because the {d-concepts} of a terminology are structured hierarchically, for a d-concept, all more general {d-concepts} can be derived as well.
	This is defined with the relation \enquote{more general than} in Chapter \ref{subsection:InheritanceConcepts}.
	The expert is usually only interested in the most specific derivable diagnosis ({d-concept}).
\end{enumerate}

\begin{Def}[\enquote{Most specific}]
	For a given $\underline{x} \in \bset$, a {d-concept} $\dconc_{spec}(\underline{x})$  is the most specific regarding  $\underline{x}$ if $\dconc_{spec}(\underline{x})$ is valid and the set of all other valid d-concepts $\dconset$ does not contain any $\dconc_{i}(\underline{x})\neq \dconc_{spec}(\underline{x})$ for which $\dconc_{i}(\underline{x}) \leq \dconc_{spec}(\underline{x})$.
	\label{Def:mostspecific}
\end{Def}

For a situation description $\underline{x} \in M$  there can be more than one \enquote{most specific} concept.
\par
To
\begin{itemize}
	\item find all valid {d-concepts} for the knowledge-based agent and
	\item visualize the most specific of them to the expert
\end{itemize}
the following approach is chosen:
The {d-concepts} in the hierarchy are attempted to be validated, starting at the root node and continuing with the children of all subsequent valid d-concepts.
If for a {d-concept} $\dconc(\underline{x})$ no children can be derived, $\dconc(\underline{x})$ is the most specific and should be the solution \cite{PGIB09}.

\subsection{Case-based reasoning}
\label{subsection:CBR}
A classic method for using knowledge gained through experience in knowledge processing is case-based reasoning (CBR) \cite{Kolo93}.

The experiences with problem solving are stored for each case.

\begin{Def}[Case]
	A case is the description of a problem situation that has already happened in real life, together with the experience that was gained during the treatment of the problem.
	The knowledge in a case base consists of a problem description $P$ and the solution $S$ as an ordered pair $(P; S)$.
	Additionally, a case can also contain explanations or an assessment $D$ of the results of the solution and thus be represented by the triple  $(P; S; D)$.
	\label{Def:Fall}
\end{Def}

\begin{figure}[!h]
	\centering
	\includegraphics[width=1.00\textwidth]{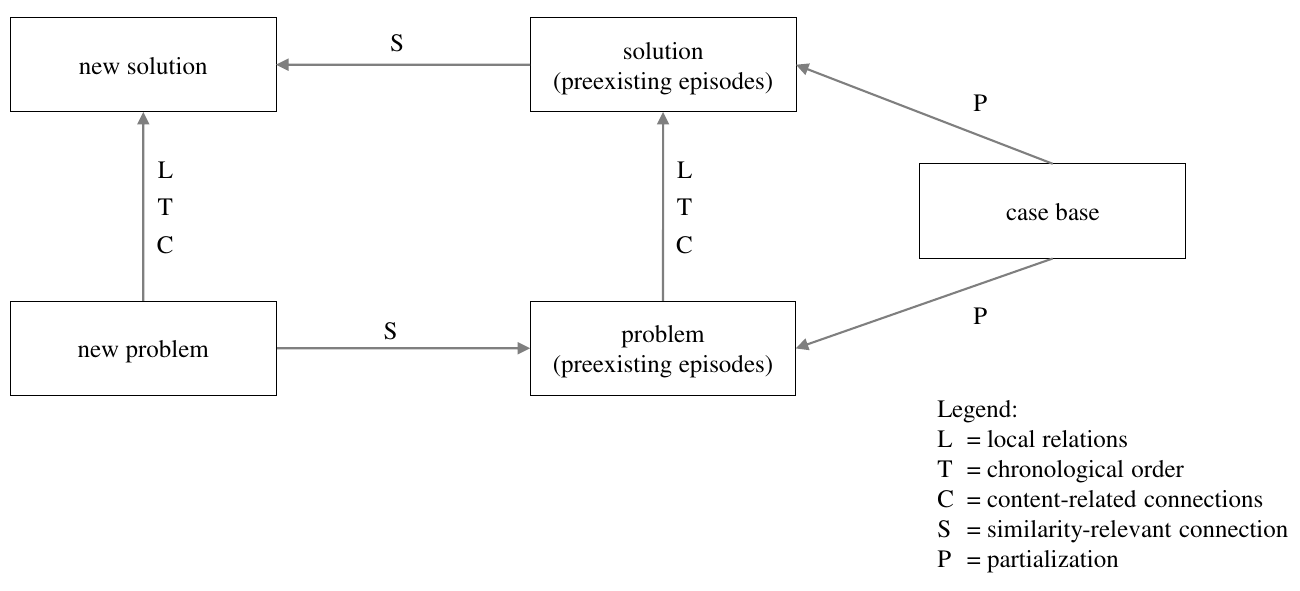}
	\caption[Case-based reasoning.]{Case-based reasoning.}
	\label{Abb:CBR}
\end{figure}

If a new problem has to be solved, a solution is searched based on previous experiences during the treatment of similar problems.
Figure \ref{Abb:CBR} illustrates the principle of reasoning.

One performance-critical aspect of CBR is the storage of resolved cases in a potentially very large case base and the efficient retrieval of cases similar to a requested case.
In previous works, the case representation often was designed specifically for the retrieval process \cite{Kolo93}.
Analogous to the initial phase of database management systems (DBMS), data independence is also desirable for a case base.
\par
The described representational paradigm uses concept hierarchies as well as instantiation and storage of cases as sequences of episodes that in turn consist of a set of term-instances. This is useful for a broad applicability of the similarity-based reasoning as well as a general infrastructure \cite{PGIB09}.
For cases with very small constraints of the available similarity measures, when just the axioms of the metric have to be fulfilled, \cite{Guhl17a} and \cite{Guhl17} describe efficient methods for the consistent storage and indexing of cases for a similarity search.

\section{Management with relational databases}
\label{sct:DB}
A knowledge-based system that only defines a representation for concepts is not sufficient.
It also has to be able to store assertional facts to infer conclusions from them.
Managing the existing data and knowledge using relational database systems has considerable benefits.
It is made possible based on the instantiation of the concept keys.

\subsection{Data storage}
A classic storage of the data in fixed data structures is not appropriate considering the potential amount of concepts, deduced concepts and instances.
Notably, access to essential operations for inference also has to be supported.
\par
The storage of instances should be as compact as possible and always represent the most specific node key.
The data set stores all valid keys.
In the normal case, the value of a key is just \enquote{true}.
The information is therefore tied to the existence of the node key.
The keys can also be used for access and references to other tables with additional information for values, texts, figures etc.
Furthermore, the respective data set contains a unique composite primary key id including the time stamp (Chapter \ref{subsection:Episode}).
\par
To specify instances and store them in databases there exist, for example, the following possibilities:
\begin{enumerate}
	\item Most specific node key
	\item Path to the most specific node key
\end{enumerate}

These will be discussed.
The structure of the storage is supposed to allow for more declarations of indexing structures, which make it possible to implement various inference operations more efficiently.

\subsection{Representation of chronological events - episodes}
\label{subsection:Episode}
Besides the representation of the terminology, concepts and  instances, it is necessary to map instances such that they describe the momentary state, a current situation $Sit_{curr}$ of the modeled section of the world.
For this purpose, episodes that depict instances of the terminological concepts and inferences at a specific moment are used.
Episodes are stored in a relational database as well.

\begin{Def}[Episode] 
	An episode describes a concrete event that happens at a certain point in time.
	\label{Def:Episode}
\end{Def}

Each episode is specified by attributes.
Every one of these attributes is defined by one or more instantiated semantic keys.
Depending on the attribute, either exactly one or multiple forms can be assigned to an episode. 
With episodic knowledge, developments over time can be represented.
In general, an episode is more closely defined through time ($T$), content ($C$) and localization ($L$) \cite{OePe98}.

\subsection{Storage of instances}
The storage of an instance should be as compact as possible and represent accessing the most specific known d-concept or concept key.
\par
The derivation of more general d-concepts or concept keys should preferably happen in constant time.
The structure should allow index structures that make it possible to efficiently find, for example, all instances that include a certain concept key.

\subsubsection{Unique keys, efficiently determining predecessor nodes}
Every node is given a unique key.

Only the key of constant length is stored, which makes this approach efficient storage-wise.
The derivation of predecessor nodes is not trivial, but the index structure can be generated.
For this, every node of a tree must store a reference to the parent node.
This can easily be achieved with a hash table or dictionary.
There, it can be computed in logarithmic time using parent pointers whether, for example, a d-concept is a predecessor of an instance.
\par
The primary benefit of this version is that the maintenance of the knowledge base, including insertions, renaming (not the indices), deletions and moving of nodes is relatively unproblematic.

\subsubsection{Identification of a node with its path}
It is also possible to store traversed paths.
For a valid instance, the node keys of the traversed path are stored, i.e. paths with indices.
If necessary, the respective named instance nodes can be stored for reindexing or documentation purposes.
\par
For example, efficient data structures can be constructed to identify all instances that include a concept, or for further inferences.
\par
Problematic for this form of storage are changes to the knowledge structure.
The reinterpretation or reindexing is algorithmically more complex, but has a unique solution.
Because changing the knowledge base should be a rare and well thought out operation, the resulting computing time should be manageable.

\subsection{Maintenance of knowledge bases}
Regarding the maintenance of knowledge bases it should not be assumed that they are in their forever valid form before the establishment of the knowledge-based system.
Their structure emerges because over time, experiences are collected with the knowledge base and structure as well as the knowledge inferred with it.
Thus, new nodes have to be inserted and old ones renamed to update the hierarchy.
Also, to keep the knowledge base manageable, it should be possible to remove nodes that proved to be irrelevant for the domain of discourse.
\par
The basic operations $Insert$ and $Delete$ not only influence the generic knowledge but also the stored instance knowledge.
In any case, such maintenance processes have to  preserve the information of previously stored instances.
They must be prevented from expiring by becoming completely indecipherable.
A misinterpretation of old instances through the maintenance of generic knowledge structures would be even more problematic.
\par
The operation $Delete$ removes nodes from the concept hierarchy.
Through the removal of nodes, the indexing of all nodes clearly remains correct, that is, a reindexing is not necessary.
In that regard, the operation is unproblematic.
If a large amount of nodes is deleted, a reindexing can still be useful, because it could substantially simplify the keys.
\par
The operation $Insert$ adds new nodes to the concept hierarchy.
Here, two cases have to be distinguished.

\begin{figure}[!h]
	\centering
	\includegraphics[width=0.5\textwidth]{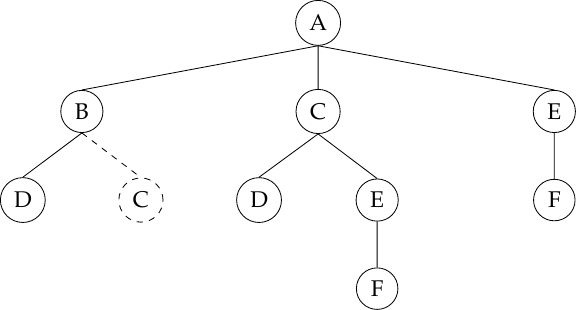}
	\caption[Insert-operation.]{Insert-operation.}
	\label{Insert}
\end{figure}

\paragraph{Already existing concept}
During the $Insert$-operation, all concepts that are more general than the concept of the newly inserted node can be kept, but all more specific concepts have to be reindexed.
Thus, $Insert$ is equivalent to the continuation of the indexing algorithm after all more general concepts have already been indexed.
In Figure \ref{Insert}, the dashed node with the concept $C$ is introduced into the existing hierarchy.
The concepts $A$ and $B$ are more general than $C$ and do not have to be reindexed.
The concepts $D$, $E$ and $F$ are more specific and have to be reindexed (together with $C$ itself).

\paragraph{New concept}
A special case appears when one or more nodes with a new concept are inserted.
Then, only these nodes have to be newly indexed.
Intuitively, the complexity of the changes during the $Insert$-Operation depends on how many concepts are affected.
\par
These operations influence the stored instance knowledge differently depending on where they are relative to a node key.
In any case, during the altering of a node (renaming or attaching to another parent node) by a knowledge engineer it has to be decided whether instances of this node will carry the same meaning after the change or if this is more akin to a deletion and reinsertion.
The same decision has to be made for all child nodes.

\subsection{Checkup modeling and knowledge base}
Important for successful work with a knowledge-based system is the modeling and formalisation of the knowledge base. Formalisation involves transforming the natural language description into a formal language description. Aspects of formalisation are
\begin{itemize}
\item semantic \enquote{normalisation}
\item syntactic \enquote{normalisation}
\end{itemize}

Clear semantics cannot be created by simply converting a vague or ambiguous or convoluted natural language sentence into a clear sentence. On the basis of the methods introduced here, some tests and, if necessary, corrections can be carried out to ensure certainty.

\begin{enumerate}
\item Completeness: 
\label{item:complete}\newline
The proof of completeness from proposition \ref{SatzVollstaendigkeit} (section \ref{subsct:Proofcompleteness}) can also be rewritten as a test of whether the given tree for indexing satisfies the conditions for \enquote{completeness}. Thus, the use of the completeness test for the knowledge engineer also brings concrete hints for the improvement of the tree and a more precise expression. If the test comes to the result \enquote{no term hierarchy}, a change, renaming or insertion of a new term at the correct level must be made to the graph.

\item Concept hierarchies, indexing of concept keys:
\label{item:Concept}
\newline			
If indexing produces concept keys with many variables, this indicates inaccurate modeling of the given knowledge base with the risk of later semantically incorrect captures, stores and processing. These indices are created when forced hierarchies of independent aspects are designed and used at different levels and on different paths in order to use as few, more general terms as possible. This also indicates that different aspects are being drawn together (see \ref{itemm:multiaxial}. Multiaxial modeling), which also leads to semantic inaccuracies and is likely to blur the fact gathering, make the knowledge base poorly maintainable and make inferences unnecessarily difficult.  

The aim should be to build very clear concept hierarchies so that indices are generated with as few variables as possible. When branching at a node with clear semantics (at a certain level), the specialising node names should be chosen adequately and specifically for the branching of these nodes and not be reused at other paths with different semantics, but then the corresponding attached specific terms there.

\item Multiaxial modeling:
\label{itemm:multiaxial}\newline
Different aspects should be represented as precisely as possible in independent subtrees. If different aspects are combined with regard to certain nodes, a multiaxial modeling of the concept hierarchy should be carried out at these positions (section \ref{subsct:multiaxial}).

\item Procedure - more differentiated hierarchical structuring of knowledge:
\newline
In order to clarify the problems outlined in \ref{item:complete}. to \ref{itemm:multiaxial}. the following procedure is recommended.

As always, the hierarchies are represented as a graph. The described graph algorithms are still used.

A possible approach for the violation of completeness is:
\begin{itemize}
	\item The original graph is joined with the subtree at a node of the tree where completeness is violated.
	\item The subtree is revised and clearly modeled according to the semantics to be displayed.
	\item The semantics, terminology and hierarchy of the new subtree created in this way must be organized or supplemented accordingly so that it satisfies the completeness test. 
	\item The subtree is indexed independently using semantic indexing algorithm according to the resulting more adequate hierarchy. The non-use of already assigned keys must be defined as an additional constraint when indexing the subtree.  
	\item The newly created subtree is attached to a suitable node of the original graph, whereby completeness must be guaranteed.  
\end{itemize}
\end{enumerate}

\section{Medical system as an example application}

In collaboration with physicians from the interdisciplinary work group \enquote{Schmerzmedizin} (pain medicine) of the DGSS (German pain society), the knowledge-based system \textit{iSuite} was developed.
It is in use since 2000 and has been updated continuously.
The system is an information and support system for the complex area of pain medicine. 
\textit{iSuite} consists of several components that access a shared knowledge base which contains generic medical knowledge as well as a case base with real patient data.
The complete knowledge is represented with a relational database system in a knowledge base.
\par
The system provides assistance to the physician in the form of automated anamnesis dialogs, documentations, research, calculations, evaluations, illustrations and suggestions during the treatment of a patient.
The system contains different components.
These components access the knowledge represented in the database.

\begin{figure}[!h]
	\centering
	\includegraphics[width=1.00\textwidth]{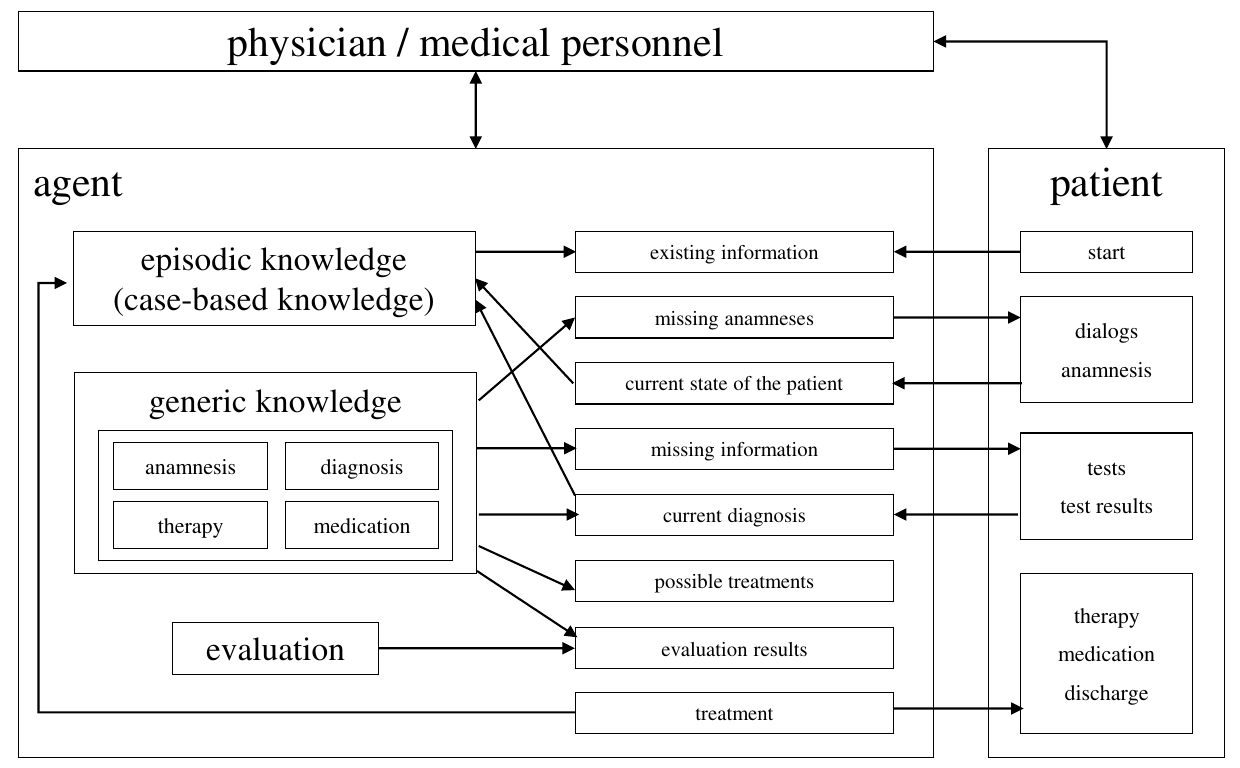}
	\caption[Medical knowledge-based agent.]{Medical knowledge-based agent.}
	\label{Abb:MedBeh}
\end{figure}

\paragraph{Knowledge-based agent}
The system architecture as a knowledge-based agent is depicted in Figure \ref{Abb:MedBeh}.

In the dialog, the agent deals with two groups of people: physicians/medical personnel and patients.
Internally the agent tries to support the steps that are necessary for the medical treatment cycle.
The knowledge used for this purpose is generally structured along two axes.
First the knowledge can be mapped to one or more treatment steps,
i.e.\ there exists knowledge about anamnesis, diagnosis, therapy and medication phase as well as connecting knowledge, e.g.\ which diagnoses were made because of a certain anamnesis.
Along a second axis, instance knowledge, e.g.\ the anamnesis of a certain patient at a certain point in time, and generic knowledge, i.e. general knowledge about the anamnesis, are distinguished.
For instance knowledge, an interaction with the patient is necessary, which happens through a patient dialog system for the gathering of anamnesis information.
The dialog system enables the periodical surveying of information necessary for diagnosis and quality control.
It also has the goal of minimizing the necessary workload.
Information on tests, diagnoses, therapy evaluations as well as their results are supplied to the agent by personnel.
All knowledge is stored in a shared generic knowledge representation (Chapter \ref{subsectionGrundbegriffe}).
\paragraph{Knowledge base}
The knowledge base for anamnesis, diagnosis, therapy and medication consists of:
\begin{enumerate}
	\item Concept terminology of the domain: \\
	The concept terminology consists of taxonomy graphs (trees) of the basic concepts and their specifications.
	The concept meaning is represented through the respective nodes and the position in the graph.
	\begin{itemize}
		\item  $K^{B}$ are lexical entries (nodes) for terminological concepts.
		\item $KM^{B}$ is the set of lexical entries for terminological concepts.
		\item $R^{B}$ is the set of relations.
		The relations are not represented explicitly but implicitly with computed concept keys for each node in the graph.
		A relation exists if nodes are partially unifiable.
		\item $R^{B} \subseteqq KM^{B} \times KM^{B}$ is a directed relation that describes the generalization and specialization relationships.
	\end{itemize}	
\end{enumerate}	

\begin{enumerate}
	\setcounter{enumi}{1}
	\item Terminology of deduced concepts: \\
	Concept terminologies represent the deduced concepts in the domain.
	A taxonomy of the deduced concepts expands the semantic expression possibilities.
	\begin{itemize}
		\item $K^{C}$ are lexical entries (nodes) for deduced concepts.
		\item $KM^{C}$ is the set of deduced concepts.
		\item $R^{C}$ is the set of relations.
		The relations are not represented explicitly but implicitly  with computed concept keys for each node in the graph.
		A relation exists if nodes are partially unifiable.
		\item $R^{B} \subseteqq KM^{B} \times KM^{B}$ is a directed relation that describes the relationships \enquote{more general than} and \enquote{more specific than}.
	\end{itemize}	 
\end{enumerate}

\begin{enumerate}
	\setcounter{enumi}{2}
	\item Deduced concept descriptions: \\
	Each deduced concept has a respective deduced concept description.
	These descriptions declare the necessary and sufficient conditions for deduced concept instances.
	Through inheritance, descriptions of superordinate deduced concepts can be passed on to subordinate deduced concepts.
	A \enquote{more general than} and \enquote{more specific than} relation is used.
	The deduced concept descriptions enable the inference process.
	\begin{itemize}
		\item $KM^{B}_{T}$ are subsets of lexical entries for concepts.
		\item $KM^{C}_{T}$ are subsets of lexical entries for references to concepts.
		\item $F \subseteqq KM^{B}_{T} \times KM^{C}_{T}$ defines references between terminological concepts, deduced concepts and deduced concept descriptions.
	\end{itemize}			
\end{enumerate}	

\begin{enumerate}
	\setcounter{enumi}{3}	
	\item Semantic indexing: 
	\begin{itemize}
		\item The terminological concept hierarchy of the domain as well as the hierarchy of deduced concepts make up the basis of uniform representation and semantic indexing.
		\item To avoid suboptimal forced hierarchization and enable precise expressions, a multiaxial representation should be used.
		\item This concept hierarchy computes and assigns a unique concept key for each terminological concept and each deduced concept.
		This enables the representation of terminological relationships through partial unification.
		\item At the same time, the concept index allows a quick search for the vocabulary of the contained concepts, their generalizations and specifications.
		The keys contain the semantically correct relations and inheritance relationships.
	\end{itemize}
\end{enumerate}

\begin{figure}[!h]
	\centering
	\includegraphics[width=0.60\textwidth]{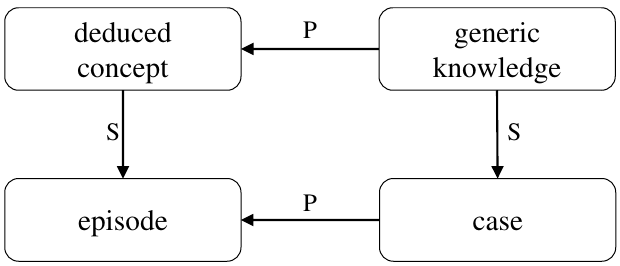}
	\caption[Abstracted structure of the knowledge base.]{Abstracted structure of the knowledge base.}
	\label{Abb:WB}
\end{figure}

\paragraph{Abstracted structure of the knowledge base}
Through the integration of concept- and case-based reasoning, specific and generic knowledge can be represented and processed uniformly.
A strongly abstracted structure of the knowledge base using partialization \textit{(P)}, specialization \textit{(S)} and similarity is shown in Figure \ref{Abb:WB}.
The partialization represents here a part-of relation, while the specialization represents an instantiation.

\paragraph{Dialog system}
On the basis of the concept terminology and the semantic indexing, a dialog system was built which supports the physician in the treatment of the patient.
The previously used representation alone is not sufficient for the dialog system because in practice additional specifications are necessary.
The concept terminology anamnesis is therefore, inter alia, expanded with:
\begin{itemize}
	\item (question-)text as a comprehensible statement
	\item alternative text for additional information, explanations, graphics, etc.
	\item question types, e.g. single-selection, multi-selection
	\item dealing with different forms of negation
	\item closed-world or open-world semantics
	\item optional, unconditional or default answers
	\item possible additional notifications
	\item etc.
\end{itemize}

The anamnesis representation has to be expanded with the respective specification, which generally can be achieved with little technical effort.
The concept terminology of the anamnesis forms the basis for the dialog system.
The terminology is processed with depth-first search and backtracking.
\par
With the use of other algorithms, the dialog can be organized more efficiently and intelligently.
It makes sense to only ask questions that are relevant for the current situation and to purposefully omit certain subtrees depending on the dialog progression and already existing knowledge.

\paragraph{Monotony and consistency conditions}
The case knowledge base should always only contain episodes that are valid, both at the current time and independently from the current answers to the remaining questions.
Additional episodes can extend the existing knowledge but not revise it.
If entries are corrected, all dependent previous answers have to be revised.
It also has to be guaranteed that the relation (more specific, more general) in Definition \ref{Def:allgspezieller} is always valid, i.e. that no other valid nodes appear below a negated node.
\par
For the knowledge base of a dialog system, for which possibly entries cannot always be interpreted unambiguously, additional generic knowledge that shows if consistency conditions have been violated can be provided.

\paragraph{Case-based reasoning}
In addition to generic domain knowledge, the practical knowledge collected in the patient base is used in the form of case-based reasoning. 
During diagnosis and therapy recommendations for a patient, similar situations to the current case are searched for in the set of all patient cases.
The recommendations given for similar previous cases are evaluated regarding their actual use as well as the resulting therapeutic success.
From these experiences, recommendations for the current case can be won or discarded.
During the therapy, success, side effects etc. are surveyed.
After the treatment of the current problem, the case becomes available in the case base as well.

\paragraph{Case base, relational database system}
During the realization of this system of CBR, multiple issues concerning demands from the case base have to be solved.
All relevant attributes of a case have to be represented and stored in the case base.
In particular, such a system is running over a long period of time and has a relatively large case base.
Therefore a data-independent consistent storage analogous to quality demands of a (relational) database management system has to be possible.
In particular, such a storage system should not be developed for one specific application.
The design of the case representation should also be generally oriented and not driven by requirements of the storage system (data independence).
In \textit{iSuite} cases are stored as sequences of episodes which in turn consist of a set of node keys.

\paragraph{Similarity measure}
In contrast to a classical relational system, similar cases regarding a similarity measure have to be efficiently found in the case base during the retrieval.
\par
Because of the size of the case base, and because the computation of the similarity function can be very complex, the approach here has to be more efficient than the naive method of searching the case base linearly.
\par
For the search for similar cases, it should first be clearly defined how similarity is measured in the context of the application.
This is a task for the domain expert of the application.
In particular, it is difficult to create a measure that corresponds to the medical similarity perception.
Furthermore, similarities in the problem domain should mirror analogies in the solution space.
For this reason, a similarity measure is usually altered multiple times during the development and use of a case-based system.
For this problem, data independence is required as well.

\paragraph{Remarks}
The idea behind the described paradigm is the equal treatment of episodic practical knowledge and general generic knowledge for an improved system performance.
To achieve this goal, different methods have to be used:
\begin{itemize}
	\item For generic knowledge, generic description logic is very powerful and has high expressiveness. Accordingly, the computations are very cost-intensive.
	\item The used calculus for application systems needs, on one hand, a high expressiveness, to be able to represent all necessary dependencies.
	On the other hand, a too generic logic should be avoided, because otherwise not enough guarantees can be given regarding the time and space complexity necessary for the computations.
	\item The described representation was developed to be able to do quick computations during run-time, e.g. for many patients and thus many episodes.
	This allows a uniform representation, effective knowledge retrieval, inferences, CBR and other necessary methods.
	\item Furthermore, the generic medical knowledge has increased vagueness, uncertainty and incompleteness. The main reasons for this are the complexity of medicine, individuality of patients and the necessity to differ from sharp descriptions.
	For \enquote{more complete} solutions to problems, apart from concept based reasoning , also Bayesian networks, which can process uncertain high-quality concepts, may be used for the generic knowledge processing.
	Here, expectation propagation is the method of choice \cite{FZP17}.
	For the determination of necessary a priori probabilities, distributions of classificatory characteristics can be obtained by the method  \cite{PDZB19}.
\end{itemize}

\section{Summary}
The indexing algorithm assigns keys which contain the complete path from the general to the most specific knowledge.
These semantic keys carry information and enable the representation of complex issues.
The described approach allows a uniform representation that on one hand permits the knowledge-based modeling and representation of practical domains in classic relational databases and on the other hand ensures an efficient analysis of the represented data and knowledge base.
It connects the two areas of knowledge representation and relational databases.
As a result of the uniform representation, the structure, architecture, and implementation of knowledge bases can be improved.
This is primarily done through the structured storage of knowledge, storing semantically clearly defined instances in a knowledge base, and the various possible inference methods.
The approach is particularly suited for domains with a clear terminological structure.
\paragraph{Acknowledgment}
We thank Mr. H.L. Biskupski for his help with the preparation of this manuscript.

\bibliographystyle{abbrv}

\bibliography{literatur}
\end{document}